\pdfoutput=1

\documentclass[11pt]{article}

\usepackage[final]{acl}
\usepackage{multirow} 
\usepackage{times}
\usepackage{latexsym}

\hypersetup{
  colorlinks=true,
  linkcolor=blue,
  urlcolor=blue,
  citecolor=blue
}

\usepackage[T1]{fontenc}
\usepackage[utf8]{inputenc} 
\usepackage{listings}
\usepackage{tabularx, threeparttable, booktabs, enumitem}
\setlist[itemize]{leftmargin=*, itemsep=1pt, topsep=0pt}
\renewcommand{\arraystretch}{1.05}

\usepackage{refcount}   
\renewcommand{\footref}[1]{%
  \hyperref[#1]{\textsuperscript{\getrefnumber{#1}}}}

\usepackage{xcolor}  

\usepackage[most]{tcolorbox}
\newtcolorbox{AARcommentbox}{
    colback=gray!10,        
    colframe=red!50,       
    boxrule=0.5pt,          
    arc=2mm,                
    left=2mm, right=2mm, top=1mm, bottom=1mm,  
    fontupper=\normalfont,  
    before upper=\parindent15pt\parskip5pt,   
    title = AAR Comment
}

\usepackage{graphicx} 
\usepackage{subcaption}
\usepackage{enumitem}
\usepackage{tabularx}
\usepackage{amsmath}

\usepackage[T1]{fontenc}

\usepackage[utf8]{inputenc}

\usepackage{microtype}

\usepackage{inconsolata}

\title{Empathy Applicability Modeling for General Health Queries}

 \author{
  Shan Randhawa\textsuperscript{1} \quad
  Agha Ali Raza\textsuperscript{2} \quad
  Kentaro Toyama\textsuperscript{1} \quad
  Julie Hui\textsuperscript{1} \quad
  Mustafa Naseem\textsuperscript{1} \\
  \textsuperscript{1}University of Michigan \quad
  \textsuperscript{2}Lahore University of Management Sciences \\
  \texttt{\{shanmr,toyama,juliehui,mnaseem\}@umich.edu} \quad
  \texttt{agha.ali.raza@lums.edu.pk}
}

\begin{document}
\maketitle
\begin{abstract}

LLMs are increasingly being integrated into clinical workflows, yet they often lack clinical empathy, an essential aspect of effective doctor–patient communication. Existing NLP frameworks focus on reactively labeling empathy in doctors' responses but offer limited support for anticipatory modeling of empathy needs, especially in general health queries.
We introduce the Empathy Applicability Framework (EAF), a theory-driven approach that classifies patient queries in terms of the applicability of emotional reactions and interpretations, based on clinical, contextual, and linguistic cues. We release a benchmark of real patient queries, dual-annotated by human annotators and GPT-4o. In the subset with human consensus, we also observe substantial human–GPT alignment. 
To validate EAF, we train classifiers on human-labeled and GPT-only annotations to predict empathy applicability, achieving strong performance and outperforming the heuristic and zero-shot LLM baselines. Error analysis highlights persistent challenges: implicit distress, clinical-severity ambiguity, and contextual hardship, underscoring the need for multi-annotator modeling, clinician-in-the-loop calibration, and culturally diverse annotation.
EAF provides a framework for identifying empathy needs \textit{before} response generation, establishes a benchmark for anticipatory empathy modeling, and enables supporting empathetic communication in asynchronous healthcare.

\end{abstract}

\section{Introduction}

Clinical empathy integrates cognitive (understanding), emotional (resonating), and action-oriented (expressing) components 
\citep{guidi2021empathy}. It is indispensable for clinical care, deepening therapeutic relationships and improving outcomes such as patient satisfaction, care effectiveness, reduced distress, and hospital length of stay \cite{guidi2021empathy, olson1995relationships, hoffstadt2020patients}; yet clinicians miss 70--90\% of empathic opportunities during patient interactions \cite{morse2008missed, hsu2012providing}.

Large Language Models (LLMs) are increasingly integrated into healthcare workflows and patient interactions, with major Electronic Health Record vendors such as Epic adopting them for clinical messaging and nearly half of physicians reporting patients consult ChatGPT before visits \cite{antoniak2024nlp, sermo2025poll}. While these trends highlight rapid adoption of LLMs in healthcare, they also raise concerns of lacking empathy crucial for asynchronous physician-patient interactions \citep{koranteng2023empathy}. However, effective empathy requires discernment, not just fluency. This highlights a critical, antecedent challenge: How can we systematically model the applicability of empathy, allowing systems to recognize the specific clinical and linguistic cues that warrant an emotional response?



Modeling empathy in text is inherently difficult without non-verbal cues, and NLP research has historically over-weighted emotional aspects while overlooking cognitive empathy \cite{lahnala2022critical}. Existing frameworks capture empathy's multidimensionality but assess it post hoc in responses, while clinical discourse work addresses anticipation but remains need-blind and tied to multi-turn synchronous settings (see Section~\ref{sec:related}). No prior work models anticipatory empathy applicability in asynchronous single-turn general health queries, which would help providers and LLMs better address empathic needs in these settings.





To address this gap, we propose the Empathy Applicability Framework (EAF), a theoretically grounded method to proactively identify when and what type of clinical empathy is warranted in response to patient queries. EAF operationalizes empathy along two key dimensions: affective (emotional reactions) and cognitive (interpretations), labeling each as Applicable or Not Applicable based on clinical, contextual, and linguistic cues within patient queries. A Not Applicable label on both dimensions signals that the patient's needs are primarily informational \cite{chai2019predicting}, and that affective and interpretive empathy would be misplaced, allowing the response to maintain a factual orientation. This binary framing is intentionally scoped to the antecedent applicability judgment, whether empathy is warranted at all, while graded need levels and uncertainty estimates tied to cue strength and ambiguity remain a natural extension for future work.

We make three primary contributions: (i) \textbf{Framework Design}: we introduce and theoretically ground the EAF in clinical empathy literature, clearly differentiating our anticipatory applicability model from prior approaches; (ii) \textbf{Annotated and analyzed Benchmark}: a novel dataset of 1,296 patient queries annotated by humans and GPT-4o\footnote{\label{fn:repo} Benchmark dataset, annotation scripts, training code, and model outputs are available at \url{https://github.com/shanmrandhawa/Empathy-Applicability-Framework}.}, demonstrating EAF’s reliability and interpretability; and (iii) \textbf{Operationalization Challenges}: we identify and systematically analyze specific contexts where anticipatory empathy annotations diverge, highlighting opportunities for future research in multi-annotator modeling, clinician-in-the-loop systems, and culturally sensitive annotation strategies.

\section{Related Work}
\label{sec:related}

Empathetic language technologies have largely been studied as response-generation problems in settings where emotional support is presumed. EmpatheticDialogues introduced an open-domain benchmark grounded in emotional situations \cite{rashkin2019towards}, catalyzing work that implicitly treats empathy as universally warranted. ESConv extends this line to multi-turn emotional support with strategy annotations rooted in Helping Skills Theory \cite{liu2021towards}. To improve response quality, cause-aware models inject emotion-cause reasoning via explicit cause recognition \cite{gao2021improving}, commonsense-augmented generation \cite{sabour2022cem}, and chain-of-thought cause-aware prompting \cite{chen2024cause}. Most recently, \citet{lee2025heart} annotate empathy-cause text spans in speaker posts, identifying which parts of the query evoke empathy from a responder, and combine these with figurative-language signals to improve generation in a mental-health support domain. More broadly, anticipatory methods incorporate foresight into empathetic generation, for example by anticipating the next dialogue development via commonsense inference \cite{wang2025sibyl}. Collectively, these approaches optimize \textit{how} to empathize but operate under the assumption that empathy is \textit{always} warranted, without modeling whether it is clinically appropriate to withhold it.

The field has also developed richer representations and evaluations of empathy in text. EPITOME operationalizes expressed empathy via three communicative mechanisms (emotional reactions, interpretations, and explorations), enabling post-hoc scoring and rationale extraction in mental-health peer support \cite{sharma2020computational}. \citet{chai2019predicting} address empathy's multidimensionality from a different angle, classifying online support responses as Informational or Emotional and linking these to downstream mental health outcomes. However, both assess empathy post hoc in the response itself, offering no guidance while a clinician is composing a reply to a patient query. In clinical settings, \citet{lahnala2024appraisal} formalize empathic opportunities and clinician elicitation/response as functions of affect, judgment, and appreciation in breaking-bad-news dialogues. This discourse-analysis lens excels at characterizing stance shifts over multi-turn synchronous conversations, yet it classifies stance, not what the patient needs (cognitive clarification vs.\ emotional warmth), remaining need-blind and unsuited to single-turn, asynchronous general health queries.

Our work differs from this body of research in three key ways. First, \textbf{emotion presence vs.\ empathy applicability}: frameworks such as EPITOME \cite{sharma2020computational}, empathy direction detection \citep{hosseini2021takes}, and empathy intent classification \cite{jiang2023empathy} detect affective states or label empathy in existing responses. EAF instead models whether empathy is warranted in a clinical context; factual uncertainty about cancer prognosis may warrant interpretive acknowledgment even without explicit affect. Second, \textbf{cause-aware enrichment vs.\ anticipatory decision}: cause-aware methods, including empathy-cause span annotation \cite{lee2025heart}, and anticipatory generation \cite{wang2025sibyl,gao2021improving,sabour2022cem,chen2024cause} optimize responses once empathy is assumed relevant. EAF performs an upstream applicability decision \textit{before} generation, complementing rather than replacing these methods. Third, unlike general-domain benchmarks \cite{rashkin2019towards,liu2021towards}, \textbf{EAF is grounded in Patient-Centered Communication functions} \cite{epstein2007patient,mccormack2011measuring}, operationalizing empathy along affective and cognitive dimensions to model clinical appropriateness rather than affect intensity alone.

\renewcommand{\arraystretch}{1.05}   
\setlength\tabcolsep{3pt}            

\begin{table*}[h]
\small
\centering
\begin{tabularx}{\textwidth}{p{2.8cm} X X}
\toprule
\textbf{Dimensions} & \textbf{Applicable cues} & \textbf{Not Applicable cues}\\
\midrule
\multirow[t]{2}{2.7cm}{%
  \raggedright\textbf{Emotional Reactions}\\[1pt]
  \footnotesize Expressions of warmth, compassion, concern, or similar feelings conveyed by a doctor in response to a patient's query.
}
& \begin{itemize}[leftmargin=*, itemsep=0.5pt]
    \item \textbf{Severe Negative Emotion}
    \item \textbf{Inferred Negative State}
    \item \textbf{Seriousness of Symptoms}
    \item \textbf{Concern for Relations}
  \end{itemize}
  \emph{Rationale:} Signals reflect distinct pathways of emotional distress, guiding when emotional reactions are warranted.
& \begin{itemize}[leftmargin=*, itemsep=0.5pt]
    \item \textbf{Routine Health Management} 
    \item \textbf{Purely Factual Medical Queries}
    \item \textbf{Neutral Symptom Descriptions}
    \item \textbf{Hypothetical Queries}
  \end{itemize}
  \emph{Rationale:} Signals no emotional content; omit reactions to maintain factual medical focus.\\[4pt]
\midrule
\multirow[t]{2}{2.8cm}{%
  \raggedright\textbf{Interpretations}\\[1pt]
  \footnotesize Communication of an understanding of the patient’s feelings (expressed or implied) and/or experiences (including contextual factors) inferred from the patient's query.}
& \begin{itemize}[leftmargin=*, itemsep=1pt]
    \item \textbf{Expression of Feeling}
    \item \textbf{Experiences or Context Affecting Emotional State}
    \item \textbf{Symptoms with an Emotional Impact}
    \item \textbf{Distressing Uncertainty About Health}
  \end{itemize}
  \emph{Rationale:} Signals lived burden, context, or uncertainty requiring interpretive acknowledgment.
& \begin{itemize}[leftmargin=*, itemsep=1pt]
        \item \textbf{Emotional‑Reactions N/A cues +: } with absence of distressing contextual
or experiential details.
  \end{itemize}
  \emph{Rationale:} Signals absence of both emotional and contextual cues, preventing over-empathizing and maintaining focus on informational needs.\\[4pt]
\bottomrule
\end{tabularx}

\caption{Empathy Applicability Framework (EAF). Each dimension lists cues for when an empathic dimension is \emph{Applicable} or \emph{Not Applicable}; brief rationales explaining what each cue set captures follow the cues. Detailed description of the EAF and its cues with examples is provided in Appendix~\ref{appendix:framework_details}. Also, see Appendix~\ref{sec:appendix-scenarios}, Table~\ref{tab:empathy_scenarios} for concrete query scenarios illustrating cues usage and EAF operationalization, including example responses demonstrating how applicability judgments shape empathetic realization.}
\label{tab:eaf_table}
\end{table*}

\section{Empathy Applicability Framework and Theoretical Grounding}

The EAF identifies empathic needs proactively by assessing patient queries along two dimensions adapted from EPITOME~\cite{sharma2020computational} and informed by Chai et al.’s distinction between emotional and informational support~\cite{chai2019predicting}: \textit{Emotional Reactions} and \textit{Interpretations}. We intentionally exclude EPITOME's third dimension, Explorations, which concerns follow-up probing and dialogic expansion in multi-turn interactions. In our single-turn, asynchronous setting, the clinician must first determine whether emotional acknowledgment or interpretive understanding is warranted before further exploration. Our framework models this antecedent applicability decision, which precedes any conversational elaboration. Table~\ref{tab:eaf_table} summarizes the EAF, detailing applicable and non-applicable cues for each dimension.

To develop EAF, we performed inductive thematic coding on 300 randomly selected patient queries from the HealthCareMagic and iCliniq datasets~\cite{li2023chatdoctor}, identified themes, formed subcategories (cues), and iteratively refined them to comprehensively and distinctly capture empathy applicability.

Additionally, we ground EAF cues in Patient‑Centered Communication (PCC) functions \cite{epstein2007patient}, to ensure their alignment with clinically valid expressions of empathy. Specifically, PCC's \textit{Responding to Emotions} function, particularly the \textit{Exploring and Identifying Emotions} domain \citep{mccormack2011measuring}, is operationalized through EAF's Emotional Reaction applicability cues, which capture both explicit and implicit distress signals (e.g., Severe Negative Emotion, Inferred Negative State, Concern for Relations). PCC's \textit{emotion-validation} domain is reflected in EAF's Interpretation applicability cues, such as Expression of Feeling, where the clinician communicates understanding of the patient's affective state. PCC's \textit{Managing Uncertainty} function is represented through Interpretation cues that capture distressing uncertainty about health or treatment, emotional impact of symptoms on daily life, and contextual hardship affecting well-being. By grounding EAF cue categories in these PCC functions, the framework operationalizes patient-centered communication theoretical constructs into detectable signals within patient queries.

\section{Methods}

To determine whether EAF is reliably interpretable across a range of clinical queries and to identify any systematic challenges
, we curated a diverse dataset of health-related queries and annotated them using the EAF, employing both human annotators and an LLM. To assess whether these annotations exhibit learnable patterns, indicating the internal consistency of EAF, we trained classifiers on the EAF-labeled data. The following subsections detail the annotation and modeling procedures.

\subsection{Data Source}

We sampled 9,500 patient queries from two publicly available datasets (HealthCareMagic and iCliniq) released by \citet{li2023chatdoctor}. We sampled 4,750 queries each from HealthCareMagic (\(\approx100k\) dialogues) and iCliniq (\(\approx10k\)), to maximize linguistic and contextual diversity and avoid overfitting to a single source. 
As these datasets are publicly available and anonymized, our IRB determined that this study was exempt from human subjects review. The datasets do not carry an explicit data license; therefore, we use them exclusively for research purposes, consistent with the authors' public release, and release our de-identified EAF benchmark\footref{fn:repo} publicly. To balance rigor and cost, 1,500 of the queries were earmarked for dual annotation by humans and GPT-4o to support reliability and error analyses, while the remaining 8,000 were annotated only by GPT-4o for predictive validity testing.

\subsection{Annotation Task} \label{sec:annotation_task}
The annotation task required using EAF to label patient queries as Applicable or Not Applicable (see Table \ref{tab:eaf_table}) on two dimensions of empathy: Emotional Reactions Applicability (EA) and Interpretations Applicability (IA). Human annotators were instructed to identify at least one best-fitting subcategory per dimension to justify their labels (they mostly selected a single best-fitting subcategory). The GPT annotations listed all relevant subcategories supporting labeling decisions.

\subsubsection{Annotator Recruitment, Training and Calibration}

Due to empathy annotation subjectivity, we prioritized consistency by avoiding crowdsourcing and instead recruited and trained two annotators from Pakistan with high English proficiency: HA1, a male with a BS in Computer Science, and HA2, a female with an MS in Linguistics. We recruited two annotators via departmental channels for about a one-month engagement. 
Informed consent to use the annotated dataset to train large language models was collected from the annotators prior to the start of the annotation process. Annotators were compensated ~US\$360 (equivalent to a local monthly research salary). 
Annotators underwent three-stage training on 200 queries (50 + 50 + 100) from a subset of 1,500, with convergence meetings after each stage to clarify misunderstandings and align labeling. Training queries were excluded from later experiments. Annotators then independently labeled the remaining 1,296 queries (four dropped due to missing content) following procedures in Section~\ref{sec:annotation_task}. Annotation instructions are detailed in Appendix~\ref{appendix:annotation_instructions}.

We \textit{intentionally} employed lay annotators to capture the patient's perspective. Prior research shows that empathy is `in the eye of the beholder' (Bernardo et al., 2018), and given that the empathy levels in the clinician's response will be perceived by the patient, lay annotators whose judgments reflect the patient/recipient experience are \textit{better-suited} for this task. Additionally, prior studies show that clinicians often overlook empathic opportunities in favor of diagnostic focus (Hsu et al., 2012) and that patients’ ratings of clinicians’ empathy often diverge from clinicians’ assessments \citep{bernardo2018physicians, hermans2018differences}.

\subsubsection{GPT Annotations}

To scale the data set and enable comparison with human annotations, we used GPT-4o via the OpenAI API, prompted to act as an expert annotator using contrastive prompting \cite{gao2024customizing}. The model was given definitions of EA and IA, subcategory descriptions with examples, and labels indicating whether each subcategory was Applicable or Not Applicable. Then it returned the matching subcategories, with the format inherently indicating the applicability class (annotation scripts available in our code repository\footref{fn:repo}). Complete prompt specifications, including those with and without our framework, are included in Appendix~\ref{app:prompt-design}. 


For the 1,296 human-annotated queries, GPT-4o generated five annotation passes per query, with final labels determined by majority vote\footnote{Majority voting ensured consistency across passes. More than 94\% of queries received the same label on the first pass and as the majority vote for both empathy dimensions, indicating minimal divergence. Hence, we report evaluation metrics only with the majority-voted labels.}. For the remaining 8,000 queries, a single-pass annotation was used due to cost constraints. This yielded two subsets: 1,296 queries labeled by both humans and GPT (with majority-voted GPT labels) and 8,000 labeled solely by GPT (single-pass annotation). \textbf{Note:} Throughout the remainder of this text, all references to GPT refer specifically to GPT-4o.

\subsection{Modeling Task and Approach}
We frame empathy applicability prediction as two independent binary classification tasks. Given a patient query \( P_i \), the objective is to predict, for each empathy dimension \( d \in \{ \text{EA}, \text{IA} \} \), whether that dimension is \textit{Applicable} (1) or \textit{Not Applicable} (0), denoted \( A_{id} \).
For each dimension, we fine-tune a distinct RoBERTa-based classifier \citep{liu2019roberta}. 
Full architectural details, including the attention mechanism, the pooling operation, and the model diagram, are provided in Appendix~\ref{appendix:model_architecture}.

\section{Evaluation Setup and Experiments} \label{sec:eval_setup}

This section details the evaluation setup and model training configurations used in our experiments.



\subsection{Annotator Agreement}
We assessed human annotation reliability using raw agreement and Cohen’s Kappa across the 1,296 independently labeled queries. For GPT-generated annotations, we compared majority-voted GPT labels with a subset of human-annotated queries: queries where both human annotators reached an agreement. This allows us to evaluate GPT performance without confounding disagreement over error or subjectivity.

\subsection{Conceptual Alignment}
To examine whether humans and GPT rely on similar rationales, we performed an UpSet plot analysis (Figure~\ref{fig:ia_ea_upset}). This analysis was limited to queries where humans and GPT agreed on the overall applicability label, allowing us to assess alignment in subcategory reasoning rather than outcome. A match is coded as \emph{Full} if GPT includes both subcategories selected by the two human annotators, \emph{Partial} if GPT's subcategories overlap with only one human's subcategory label, and \emph{No match} if GPT matches neither human subcategory.

\subsection{Divergence Bar and Qualitative Analysis}
\label{divergebaranalysis}
Given the subjective nature of empathy, we analyze mismatches as directional divergences rather than strict errors. To characterize disagreement, we use three-way divergence bars (Figure~\ref{fig:divergenceBars}) that decompose label mismatches within each subcategory into \emph{Annotator Spread} (one human labeled Applicable, the other Not), \emph{LLM-Adds} (GPT labeled Applicable, humans Not), and \emph{LLM-Omits} (GPT labeled Not, humans Applicable). Furthermore, we performed qualitative analysis on a subset of queries where GPT labeled differently, and identified thematic patterns that highlight the divergence.


\subsection{Model Evaluation}
\label{modeleval}
We evaluated the performance of the classifiers trained to predict empathy applicability (Applicable vs. Not Applicable). Reported metrics include accuracy, weighted F1 score, and macro-averaged F1 score across both dimensions (EA and IA).
To contextualize classifier performance, we compared results against four baselines: Random Guessing (assigns labels at random), Always Applicable, Always Not Applicable, and \texttt{o1}-Zero-Shot (based on OpenAI's reasoning model, without invoking the empathy applicability framework). For the \texttt{o1} baseline, we provide only the definition of the target dimension (EA or IA) and prompt it to classify each patient query as `Applicable' or `Not Applicable', preserving the zero-shot setting without framework cues. These baselines help determine whether our trained models learn meaningful patterns beyond simple heuristics or zero-shot LLM reasoning. In addition, following best practices for sanity checking and overfitting control, we include \emph{classical} text classifiers trained and evaluated on the Human Set (Section~\ref{sec:datasets}): Logistic Regression (LR) and Linear SVM with TF--IDF features (1–3 grams). These linear models provide a transparent reference for what is learnable from local lexical features.


Our goal is to evaluate whether EAF encodes structured, machine-learnable patterns rather than pursue state-of-the-art performance; for this reason, we adopt the controlled setup described above, allowing us to demonstrate that the task is learnable and not tied to a specific architecture.






\subsection{Model Training and Training Sets}\label{sec:datasets}
Each classifier for the EA and IA tasks is based on RoBERTa‑base ($\approx$125M parameters) and was trained on two distinct datasets (data and scripts available in our repository\footref{fn:repo}):
\textbf{Human Set:} Contains only queries where both human annotators reach consensus on a label for a given dimension, serving as a high-fidelity benchmark aligned with human judgment.
\textbf{Autonomous Set:} Consists of GPT-labeled data from the 8,000-query pool, with no human supervision. This tests whether models trained solely on GPT output can approximate human consensus.

For the Human Set, we split the data into subsets of training (75\%), validation (5\%), and test (20\%). For the Autonomous Set, training was done entirely on GPT-labeled data, but testing used the same human-consensus test set as the Human Set to enable consistent evaluation relative to human agreement.
Training used a single NVIDIA A40 GPU per run. A Human‑Set run finished in $\approx$15 min GPU time, while an Autonomous‑Set run took $\approx$40 min; thus the total compute budget per dimension is <1 GPU‑hour. All models were trained for 10 epochs using a learning rate of \(2\times10^{-5}\) and a batch size of 8. To ensure comparability, all models shared the same architecture and hyperparameters.


\section{Results}
In this section, we present our findings related to the reliability of the EAF and the challenges in operationalizing it. Additional dataset characterization beyond agreement and modeling results is provided in Appendix~\ref{app:dataset-analyses}.



\begin{table}[t]
\centering
\small
\setlength{\tabcolsep}{6pt}
\begin{tabular}{lcc}
\toprule
\multirow{2}{*}{\textbf{Dimension}} & \multicolumn{1}{c}{\textbf{Human–Human}} & \multicolumn{1}{c}{\textbf{Human–GPT}} \\ 
& $\kappa$ (agree / disagree) & $\kappa$ (agree/ disagree) \\ 
\midrule
EA & 0.521 (981 / 315) & 0.614 (667 / 153) \\
IA & 0.404 (898 / 398) & 0.659 (681 / 139) \\
\bottomrule
\end{tabular}
\caption{Cohen’s $\kappa$ with agreement counts: human–human agreement on the full set and human–GPT alignment on the human‑consensus subset.}
\label{tab:agreement}
\end{table}


\begin{table*}[h]
    \centering
    \small
    \caption{ Classification results across training sets and baselines (single run on the human-consensus test set). Bold indicates best performance. Classical baselines (TF--IDF+LR/SVM) are trained on the Human Set only. Our transformer significantly outperforms all the baselines. 
    }
    \label{tab:classification_results}
    \begin{tabular}{lccc|ccc}
        \hline
        \multirow{2}{*}{\textbf{Training Set / Model}} & \multicolumn{3}{c|}{\textbf{EA}} & \multicolumn{3}{c}{\textbf{IA}} \\
        & Acc & Macro-F1 & Wtd-F1 & Acc & Macro-F1 & Wtd-F1 \\
        \hline
        Random & 0.47 & 0.47 & 0.47 & 0.44 & 0.43 & 0.44 \\
        Always Applicable & 0.52 & 0.34 & 0.36 & 0.53 & 0.35 & 0.37 \\
        Always Not Applicable & 0.48 & 0.32 & 0.31 & 0.47 & 0.32 & 0.30 \\
        \texttt{o1} Zero-Shot & 0.55 & 0.40 & 0.41 & 0.62 & 0.53 & 0.54 \\
        \hline
        \multicolumn{7}{l}{\textit{Human-supervised models (train and tested on human-consensus set)}}\\
        Logistic Regression & 0.84 & 0.84 & 0.84 & 0.80 & 0.80 & 0.80 \\
        Linear SVM          & 0.83 & 0.83 & 0.83 & 0.77 & 0.77 & 0.77 \\
        \textbf{Transformer (RoBERTa-base)} & \textbf{0.92} & \textbf{0.92} & \textbf{0.92} & \textbf{0.87} & \textbf{0.87} & \textbf{0.87} \\
        \hline
        \multicolumn{7}{l}{\textit{Autonomous-supervised model (train on GPT labels, test on human-consensus test set)}}\\
        Transformer (RoBERTa-base) & 0.85 & 0.85 & 0.85 & 0.78 & 0.77 & 0.77 \\
        \hline
    \end{tabular}
\end{table*}


\begin{figure*}
  \centering

  \begin{subfigure}[b]{0.48\linewidth}
    \includegraphics[width=\linewidth]{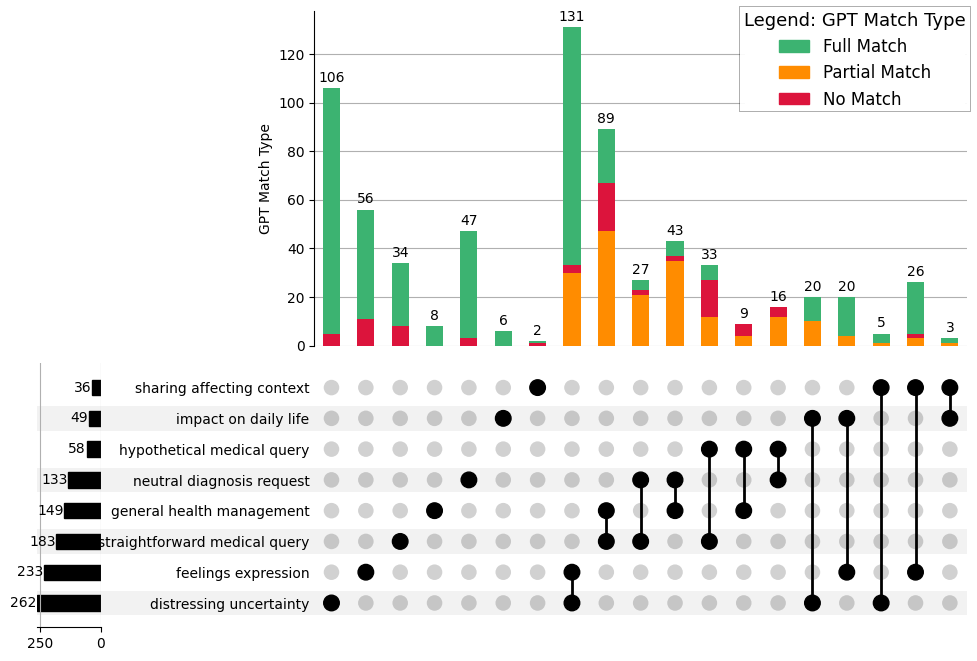}
    \caption{IA subcategory matches}
    \label{fig:upset-ia}
  \end{subfigure}
  \hfill
  \begin{subfigure}[b]{0.48\linewidth}
    \includegraphics[width=\linewidth]{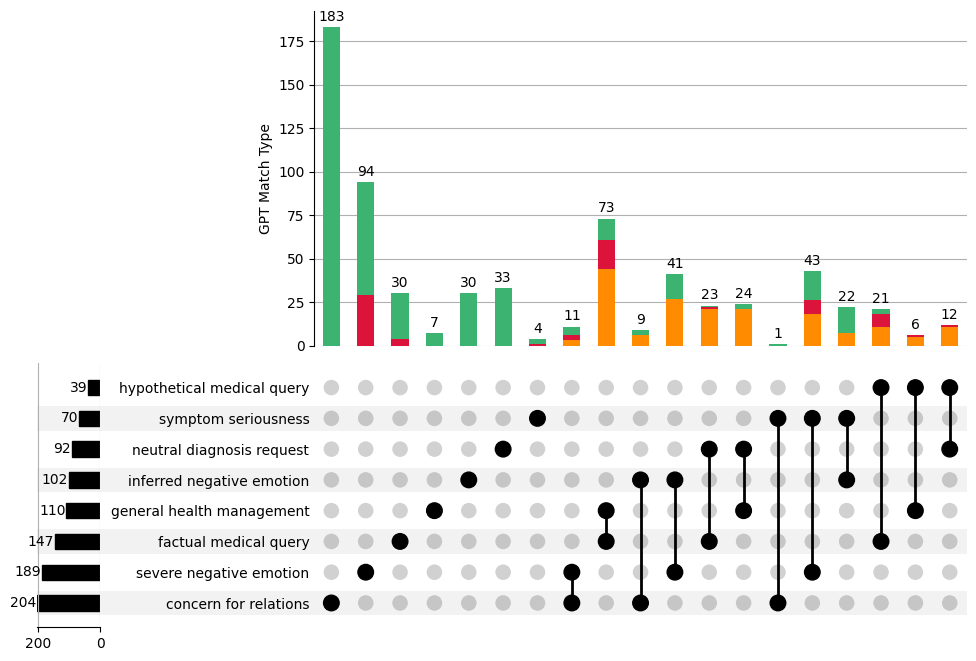}
    \caption{EA subcategory matches}
    \label{fig:upset-ea}
  \end{subfigure}


   \caption{UpSet plots comparing GPT and human rationales for \textbf{(a)} Interpretations Applicability (IA) and \textbf{(b)} Emotional Reactions Applicability (EA) subcategories. For each query, each human annotator selects one best-fit subcategory for their rationale; thus the human set is either a \textit{single-dot} combination (both humans chose the same subcategory) or a \textit{two-dot} combination (humans chose different subcategories). Horizontal bars show how often each subcategory appears in the human annotation set across all queries. Each vertical bar shows the frequency of a unique human-combination and is split by GPT agreement: \textcolor{green}{Full} (GPT’s subcategory set covers the entire human set), \textcolor{orange}{Partial} (GPT matches only one of the two human subcategories), and \textcolor{red}{No match} (GPT matches neither human subcategory). 
  }

  \label{fig:ia_ea_upset}
\end{figure*}

\subsection{Reliability of the EAF}
We evaluated reliability along three axes: Consistency, Predictive Validity, and Conceptual Alignment.


\paragraph{Consistency.}
We first assess agreement on Applicable/Not Applicable labeling between human annotators across 1,296 queries, and between GPT-4o and the \textit{human consensus} on a subset of 820\footnote{For the full set, Appendix \ref{appendix:human_gpt_agreement} (Table \ref{tab:kappa_human_gpt_scores}) shows comparable GPT agreement with each human annotator on affective EA, but substantially more variable agreement on cognitive IA.} queries. 
As shown in Table~\ref{tab:agreement}, human annotators achieved moderate agreement on both empathy dimensions, with an overall Cohen’s~$\kappa$ of 0.46. This falls within the typical range for empathy annotation tasks. \citet{kumar2026large} report that interrater reliability varies widely across empathy frameworks, with mean expert $\kappa_w$ (quadratically weighted) values of 0.41 for EmpatheticDialogues, 0.46 for EPITOME, 0.55 for Lend an Ear, and 0.60 for Perceived Empathy, indicating that moderate agreement is typical in tasks requiring inference about latent emotional states, even among expert annotators. Similarly, Sibyl \cite{wang2025sibyl} reported scores between 0.4 and 0.6. Our score is consistent with these benchmarks. Notably, agreements outnumbered disagreements \textit{by a factor of two to three}. Together, these results suggest that the EAF supports relatively consistent human labeling despite the inherent subjectivity of empathy.



GPT aligned well with the \textit{human consensus dataset} (queries where both humans agreed), achieving three-way agreement. For both EA and IA, Cohen’s~$\kappa$ exceeded 0.6 and raw agreement was about 80\% (Table \ref{tab:agreement}). These results reflect agreement
on human-aligned cases,  demonstrating
EAF’s effectiveness in guiding GPT to anticipate
empathy applicability in clearer contexts, excluding more ambiguous or complex queries (see Section \ref{challenges}).

\paragraph{Predictive Validity.}
We next evaluated whether EAF annotations are machine-learnable. As shown in Table~\ref{tab:classification_results}, classifiers trained on human consensus data achieved high performance: LR attains \textbf{0.84} Macro-F1 for EA and \textbf{0.80} for IA (SVM: 0.83/0.77), establishing a strong classical reference. Our transformer (RoBERTa-base) exceeds these values significantly \textbf{(EA: 0.92 vs.\ 0.84; IA: 0.87 vs.\ 0.80)}. Models trained on GPT-only annotations (the Autonomous set) also performed well, achieving around \textbf{0.85} for EA and \textbf{0.77} for IA on the same held-out human-consensus test set, reflecting expected loss from noisier labels or differences from human labeling. 
Our models also significantly outperformed the trivial baselines (random guessing, always applicable, always not applicable, and \texttt{o1} Zero-Shot), which yielded substantially lower scores. McNemar’s test \cite{mcnemar1947note} confirmed statistical significance of the transformers over the trivial baselines (max \(p<10^{-4}\)) and over the classical baselines (max \(p\leq 0.02\)).
Taken together, strong linear performance indicates consistent linguistic realizations of the constructs, while the transformer’s margin suggests benefits from broader context rather than overfitting. Overall, these results show that EAF-labeled data encode structured and learnable patterns.

\paragraph{Conceptual Alignment.}
We further examined whether humans and GPT rely on similar reasoning when assigning EAF labels. UpSet plot analysis (Figure~\ref{fig:ia_ea_upset}) shows strong conceptual alignment. In many cases, both human annotators independently selected the same subcategory and GPT matched it, especially for both applicability and non-applicability cues such as \textit{Severe Emotion} or \textit{Factual Queries}. These matches indicate that the EAF defines meaningful categories that are consistently identifiable by both humans and LLMs.

When annotators selected different subcategories for the same label, GPT often matched both. For example, in queries involving both \textit{Expression of Feeling} and \textit{Distressing Uncertainty}, GPT cited both reasons, suggesting that GPT can reconcile diverse human rationales and underscores the framework’s breadth in conceptualizing clinical empathy. No-match cases are rare, and GPT typically overlaps with at least one human subcategory. Appendix~\ref{app:match-miss} quantifies and corroborates these trends, showing match rates (match vs.\ miss, conditioned on humans using that subcategory) above 80\% for most subcategories.

Collectively, these results establish that EAF supports consistent human judgments, yields learnable patterns, and promotes interpretable reasoning across both humans and LLMs, making it well suited for anticipatory empathy modeling in clinical settings.


\label{RQ3EAFfalter}
\begin{figure*}[h]
  \centering
  \includegraphics[width=\linewidth]{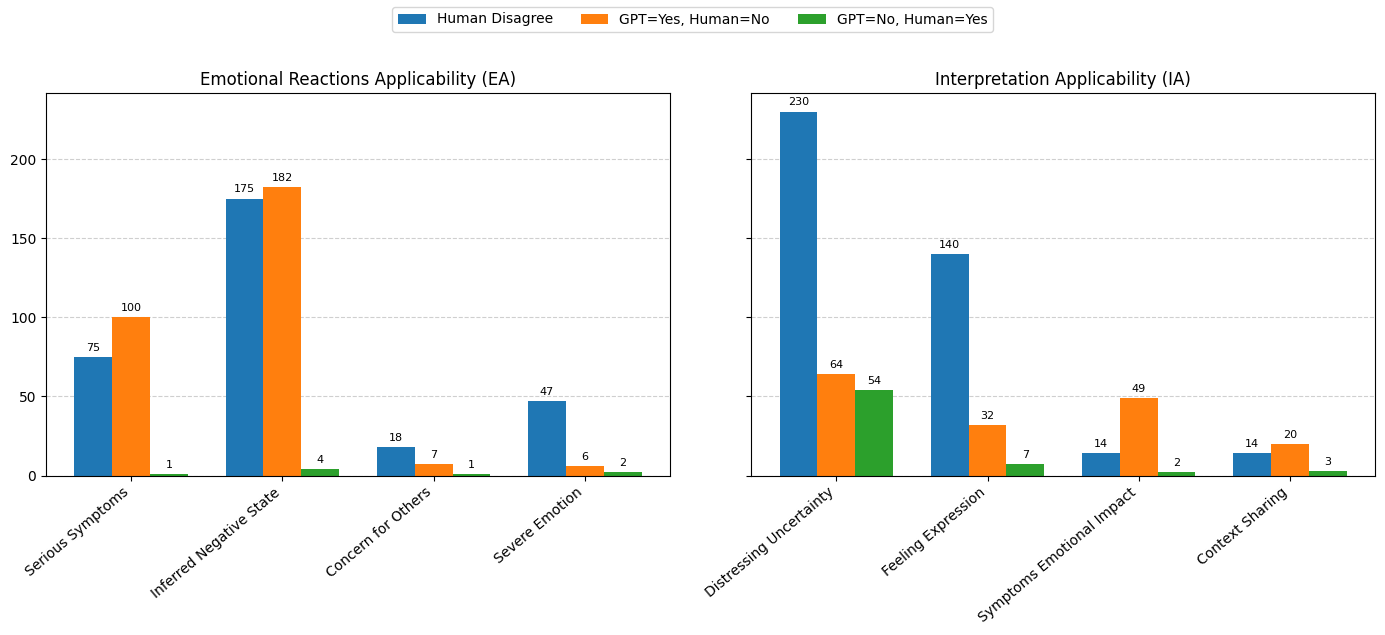}
  \caption{Three-way divergence for every subcategory.  
  Blue = \emph{Annotator Spread in Humans} (One Applicable, other not); Orange = \emph{LLM‑Adds Empathy Dimension} (GPT Applicable, Humans Not);  
  Green = \emph{LLM‑Omits Empathy Dimension} (GPT Not, Humans Applicable).  
  }
  \label{fig:divergenceBars}
\end{figure*}

\subsection{Systematic Challenges in Operationalizing Anticipatory Empathy}\label{challenges}

Divergence bar analysis (Section~\ref{divergebaranalysis}) revealed that inter-human agreement is significantly lower for IA than for EA (Table~\ref{tab:agreement}), and that despite moderate overall human-GPT agreement (Table~\ref{tab:agreement}), there is divergence at the subcategory level. Subsequent qualitative analysis  revealed three key challenges in applying the EAF, with implications for any clinical empathy framework in NLP.

\subsubsection{Challenge 1: Subjectivity in Identifying Implied Distress}\label{subjectivedistresschallenge}
The categories \emph{Inferred Negative State} (EA) and \emph{Distressing Uncertainty} (IA) show substantial divergence in inter-human and human-GPT annotations (Figure~\ref{fig:divergenceBars}). 

A qualitative review of 50 randomly selected cases\footnote{\label{fn:datafile}\textit{Detailed patient queries, mis‑aligned labels, and qualitative interpretations are available as \texttt{misalignment\_analysis.csv} at \url{https://github.com/shanmrandhawa/Empathy-Applicability-Framework}}} (25 each for Distressing Uncertainty and Inferred Negative State)\footnote{\label{fn:datafilesize}\textit{a sample size consistent with prior clinical-NLP error analyses \cite{hu2024improving}}} by the first author acting as adjudicator  revealed that in more than 50\% of the queries, one could reasonably infer implied emotional distress \textit{or} determine that the query is driven by factual intent. For instance, the female annotator labeled a pain-and-menstrual-cycle query as Distressing Uncertainty, while the male annotator treated it as a factual diagnostic request, illustrating the subjectivity of distress inference.

\subsubsection{Challenge 2: Clinical-Severity Ambiguity}
In the category \emph{Serious Symptoms} (EA), GPT labeled 100 queries as requiring emotional reactions when humans did not (Figure~\ref{fig:divergenceBars}). Qualitative analysis of 25 randomly selected cases\footref{fn:datafile} where only GPT had labeled empathy as applicable revealed three patterns: (1) In 40\% of the cases, GPT appropriately flagged empathy as needed for patients with chronic or life-threatening conditions (e.g., post-liver transplant complications) that human annotators with no medical background had overlooked; (2) borderline cases with reasonable disagreement (16\%), such as prolonged low-grade fever after kidney stones; and (3) GPT overgeneralization of vivid but non-serious pain symptoms (44\%) that did not meet the EAF criteria of chronic or life-threatening severity (for example, lip numbness after dental problems). 

\subsubsection{Challenge 3: Contextual Hardship}
GPT frequently over-applied \emph{Symptoms Emotional Impact} (SEI) and \emph{Context Sharing} (CS) tags compared to humans (Figure~\ref{fig:divergenceBars}). An analysis of 25 randomly selected\footref{fn:datafile} mismatched labels in the SEI category and all 20 mismatches in CS revealed that while GPT sometimes correctly identified complex distress signals humans missed (20--25\% of the cases), it more often (75--80\% of the cases) equated physical discomfort with emotional distress -- potentially reflecting Western-centric training biases \citep{johnson2022ghost, cao2023assessing}. 

These challenges, rooted in subjective inference, clinical ambiguity, and cultural variation, highlight the complexity of implementing clinical empathy. Addressing them requires moving beyond single-annotator consensus toward frameworks that embrace interpretive pluralism, clinical expertise, and cultural sensitivity.


\section{Discussion and Conclusion}

Asynchronous patient communication requires \emph{anticipatory} mechanisms that can signal empathic needs \emph{before} a response is written. EAF addresses this gap by assigning applicability labels to patient queries, indicating whether empathy is warranted and which dimension (emotional versus interpretive) is applicable. This framing complements cause-aware and anticipatory empathetic response generation methods (see Section \ref{sec:related}). While those approaches enrich how empathy is expressed once assumed relevant, EAF identifies the empathic needs embedded in the patient's query prior to generation. EAF can be integrated with such methods to guide empathetic response generation in clinical and general-health settings.


However, EAF faces challenges from \textbf{subjective inference}, particularly when cues are implicit (e.g., \emph{Inferred Negative Emotional State}). As appraisal theory suggests, divergent interpretations of distress often reflect genuine ambiguity rather than noise \citep{wondra2015appraisal}. Furthermore, these appraisals are shaped by \textbf{cultural priors of emotion}. \citet{eichbaum2023empathy} warn that Western-centric empathy models can misfire cross-culturally. Indeed, GPT-4o, trained on predominantly Western data \citep{johnson2022ghost}, often labeled minor inconveniences as empathy-worthy where our South Asian annotators did not. This highlights a \textbf{cultural bias} inherent to LLMs that encode American norms \citep{cao2023assessing}.

The NLP community increasingly embraces this variability through multi-annotator models and annotator-aware representations that yield calibrated uncertainty estimates and capture interpretive styles \citep{davani2022dealing, mokhberian2024capturing}. \citet{gordon2021disagreement}'s \textit{jury learning} further shows how selecting annotator subsets aligned with demographic perspectives can preserve pluralism. In clinical empathy contexts, retaining subjective variability can help anticipate diverse patient needs, including domain-specific perspectives (e.g., oncologists who prioritize emotional support as central to care) \citep{dekker2020clinical}. Building on this, we advocate extending disagreement-aware and annotator-aware frameworks toward \textbf{diversity-aware modeling} that explicitly accounts for culturally patterned differences in what counts as an empathy need, rather than collapsing them into a single consensus label.

Importantly, EAF does not assume that empathy applicability is purely objective. Rather, it provides a structured cue-based scaffold within which interpretive variability is an inherent property of clinical communication. As our results demonstrate, some cues are reliably interpreted across annotators (e.g., Severe Negative Emotion), while others involve subjectivity (e.g., Inferred Negative State) or cultural variation (e.g., contextual hardship). EAF thus models constrained inference rather than strict objectivity: cues bound the interpretive space without eliminating subjectivity. Future extensions may incorporate diversity-aware modeling without abandoning these cue-based foundations.

This work makes three contributions to clinical empathy in NLP. First, we introduce the Empathy Applicability Framework (EAF), shifting from reactive to anticipatory applicability modeling. Second, we establish a benchmark of 1,296 patient queries demonstrating reliable EAF labels. Third, our analysis identifies challenges, namely subjective inference, clinical-severity ambiguity, and contextual hardship, as opportunities to embrace interpretive pluralism via multi-annotator frameworks. By combining a practical framework with empirical operationalization, this work advances empathy modeling that respects interpretive complexity while remaining computationally tractable. All annotation scripts, training code, model outputs, and the de-identified EAF benchmark dataset are publicly available at \url{https://github.com/shanmrandhawa/Empathy-Applicability-Framework}.


\section{Limitations}

Our study faces five key constraints, the first two mirroring limitations reported by \citet{ali2025hlu}. First, we relied on only two human annotators, neither of whom had clinical training, which limited the range of perspectives represented. While lay annotators capture the patient-perspective signal and our goal was to establish the feasibility of the EAF and demonstrate moderate consistency in a cognitively complex task, two individuals cannot represent population-level variability and may introduce individual biases. Moreover, as our analysis shows, non-clinician annotators may lack the expertise to resolve cases where symptom severity affects empathy judgments. Expanding the size, clinical expertise, and cultural diversity of the annotator pool would better capture the variability of empathy judgments and reduce individual bias.


Second, all automatic annotations were produced with GPT-4o, selected for its widespread availability through ChatGPT, but this exclusive focus on the GPT series limits the generalization of our findings to other model architectures (e.g., Gemini, Claude, reasoning models, or open-source alternatives). Third, human annotators selected a single most-salient subcategory per dimension, while GPT-4o returned multiple subcategories; this procedural mismatch hinders direct comparison of disagreement patterns, and aligning the guidelines would allow for more rigorous evaluation. 


Fourth, our modeling experiments use RoBERTa-base to establish learnability rather than maximize performance. Exploring larger pre-trained models (e.g., ModernBERT), larger language models (e.g., Gemini), and prompting-based approaches could further characterize task difficulty and yield stronger classifiers, representing a natural direction for future work.

Fifth, our binary Applicable/Not Applicable framing captures the antecedent decision but does not model the intensity of empathic need. Extending EAF to incorporate graded tiers (e.g., low, moderate, high applicability) or uncertainty-based calibration could effectively capture variation in empathic need and better handle borderline cases. However, such extensions would require substantially more annotated data and careful calibration to ensure reliable predictions.

Future work should therefore involve a more diverse set of human annotators, evaluation across multiple LLM families trained under different specifications, and standardized annotation procedures between humans and models. It should also include exploration of larger pre-trained and language models using prompting-based approaches and investigation of graded empathy tiers with uncertainty-based calibration to obtain broader insights for improving empathy modeling in NLP for clinical contexts.

\section{Ethical considerations}

We developed the EAF to augment, not replace, clinician empathy judgments. Deploying EAF therefore requires close attention to several intertwined ethical risks that must be mitigated through thoughtful design and implementation.

A primary concern is the moral and social impact of artificial empathy. Because LLMs lack authentic emotional experience, we must ask whether the `applicable emotional reactions' they generate can truly convey warmth or connection. If users perceive these reactions as hollow or manipulative, an \textit{uncanny valley} effect could ensue, in which attempted comfort backfires by appearing inauthentic. Determining \textit{whether, when,} and \textit{how} automated empathy is applicable, and how to address potential deception or user discomfort, requires a systematic study of user perceptions of authenticity versus artificiality.

A second mirror-image danger arises from the same gap between simulated language and genuine feeling. As \textit{Empathic AI Can’t Get Under the Skin} discussed, LLMs lack the biological and psychological underpinnings that ground human empathy, yet their empathic language can evoke real emotional responses~\cite{NatureMI2024EmpathicAI}. \citet{kirk2024benefits} warn that users may form perceived emotional bonds with such systems, risking unhealthy attachment or disclosure of sensitive information~\cite{NatureMI2024EmpathicAI}. Thus, rejection born of perceived inauthenticity and devotion born of mistaken authenticity are twin failure modes rooted in the same ontological limitation.

For these reasons, we insist that the EAF be used strictly within a \textit{human-in-the-loop} pipeline. Clinicians must retain final authority over how and when empathy is expressed, supported by transparent rationales and safeguards that guard against both deceptive alienation and false intimacy, thus protecting patients from the dual harms of artificial empathy.



\bibliography{custom}

\appendix

\section{Empathy Applicability Framework Details}
\label{appendix:framework_details}

\subsection{Emotional Reactions in General Health Queries}

\subsubsection{Definition}
Emotional Reactions refer to expressions of warmth, compassion, concern, or similar feelings conveyed by a doctor in response to a patient's query. These reactions aim to provide emotional support and reassurance to the patient.

\subsubsection{Emotional Reactions Not Applicable}
Emotional reactions are not necessary or expected in the doctor's response when the patient's query is factual, neutral, or a simple advice request, without expressing emotional distress. Below are detailed categories reflecting when emotional reactions are not applicable:

\textbf{1. Purely Factual Medical Queries}  
\textbf{Description:} The patient requests specific medical information, including explanations of medical concepts, without emotional distress or underlying distressing uncertainty.

\textbf{Examples:}
\begin{itemize}
    \item "What is the use of Tylenol?"
    \item "Is it possible to outgrow a seafood allergy?"
\end{itemize}

\textbf{2. General Health Management Without Emotional Involvement}
\textbf{Description:} The patient seeks guidance on health management, follows up on prior advice, or requests basic guidance on minor health issues, without expressing emotional distress or underlying distressing uncertainty. Here the guidance is on what the patient should do.

\textbf{Examples:}
\begin{itemize}
    \item "I'm managing diabetes with insulin. How often should I check my blood sugar levels?"
    \item "I have swelling in my ankle after a long walk. Should I be concerned?"
    \item "I had an X-ray for a fracture; should it be strapped or cast right away?"
\end{itemize}

\textbf{3. Diagnosis Requests with Neutral Symptom Descriptions}
\textbf{Description:} The patient describes symptoms neutrally without expressing emotional distress or underlying distressing uncertainty. Here the request is about asking what the doctor thinks the issue is.

\textbf{Examples:}
\begin{itemize}
    \item "I have intermittent knee pain from working out. How would I know if I tore cartilage?"
    \item "Hello. I am having pain in my jaw area, immediately in front of my left ear. The pain is random. My feeling is it is somehow related to sinus but that's just a gut feeling."
\end{itemize}

\textbf{4. Hypothetical Medical Queries Without Emotional Concern}
\textbf{Description:} The patient inquires about hypothetical situations without emotional involvement.

\textbf{Examples:}
\begin{itemize}
    \item "If someone has XYZ symptoms, what might be the cause?"
    \item "What would happen if a person skipped their medication?"
\end{itemize}

\subsubsection{Emotional Reactions Applicable}
\textbf{Definition:} Emotional reactions are necessary or expected in the doctor's response when:
\begin{itemize}
    \item The patient expresses emotions like fear, worry, frustration, or distress.
    \item The patient implies emotional distress over symptoms affecting their well-being.
    \item The patient's tone suggests a need for reassurance or emotional support.
    \item The patient is expressing concern for a close relation (e.g., a child, spouse).
\end{itemize}

Below are detailed categories reflecting when emotional reactions are applicable:

\textbf{1. Seriousness of Symptoms}
\textbf{Definition:} The patient describes symptoms that suggest a life-threatening or chronic health condition significantly impacting long-term health or quality of life. This includes diseases like cancer, heart disease, mental health issues, or chronic conditions leading to disability. The symptoms suggest a life-threatening or serious health condition that could significantly impact long-term health or quality of life.

\textbf{Examples:}
\begin{itemize}
    \item "My father has been having severe chest pains and shortness of breath. Could it be a heart attack?"
    \item "I've been experiencing numbness and weakness in my limbs for months. Could this be multiple sclerosis?"
    \item "I'm 78 and have been told I have a floating hernia after bowel cancer surgery. Can it be cured?"
\end{itemize}

\textbf{2. Severe Negative Emotion Expressed}
\textbf{Definition:} The patient explicitly states intense emotions such as fear, frustration, or anger regarding their health.

\textbf{Examples:}
\begin{itemize}
    \item "I feel depressed and anxious like never before. I cannot sleep at night."
    \item "I am scared and plan on taking my son to the doctor. Should I be overly worried?"
    \item "I'm terrified about my recent diagnosis of cancer."
\end{itemize}

\paragraph{3. Underlying Negative Emotional State Inferred}

\textbf{Definition:} The patient implies emotional distress that isn't explicitly stated but can be inferred from their tone or descriptions, such as subtle signs of emotional worry, frustration, or distress about delays or uncertainties. Focus on emotional worry, not the medical concern.

\textbf{Examples:}
\begin{itemize}
    \item "I am starting to get a little alarmed by this spotting after ovulation. Is this cause for concern?" (Worry inferred)
    \item "I have been trying to conceive, and the report does not look right to me. I just want to take a second opinion." (Anxiety inferred)
    \item "I need to be a bit more at ease after what I read about diabetic enteropathy. I was a bit scared if it might be fatal." (Fear inferred)
\end{itemize}

\textbf{4. Concern Severity for Close Relations}
\textbf{Definition:} The patient is asking on behalf of someone with whom they share a close, protective relationship, implying heightened emotional concern.

\textbf{Examples:}
\begin{itemize}
    \item "Hello, I am the mother of a five-year-old. He has a small lump that hasn't gone away. Should I take him to a dermatologist?"
    \item "My son recently started daycare and has gotten sick. His fever was 102.9. Should I take him to the hospital?"
\end{itemize}

\subsection{Interpretations in General Health Queries}

\subsubsection{Definition}
Interpretations refer to the communication of an understanding of the patient’s feelings (expressed or implied) and/or experiences (contextual factors) inferred from the patient's query. It's about recognizing and articulating what the patient is feeling and why, based on their situation, concerns, and history. 

\subsubsection{Interpretations Applicable}
Interpretations are necessary when the patient's query requires the doctor to communicate an understanding of the patient's feelings (expressed or implied) and/or experiences (contextual factors). This involves acknowledging emotions, underlying concerns, or contextual elements that influence the patient's emotional state. Below are detailed categories reflecting when interpretations are applicable:

\textbf{1. Expression of Feelings (Explicit or Implicit)}

\textbf{Description:}  

The patient expresses emotions directly or implies them through language or tone. This includes feelings such as fear, anxiety, frustration, sadness, or hopelessness.

\textbf{Examples:}
\begin{itemize}
    \item \textbf{Explicit Expression:} 
    \begin{itemize}
        \item "I'm really scared about these chest pains."
        \item "I'm frustrated because my symptoms aren't improving."
        \item "I have been in severe pain. It hurts so bad getting out of bed."
    \end{itemize}
    
    \item \textbf{Implicit Expression:}
    \begin{itemize}
        \item "I guess I have to accept this is how things will be now."
        \item "Nothing seems to be helping."
        \item "I don't know what to do anymore."
    \end{itemize}
\end{itemize}

\textbf{2. Sharing Experiences or Contextual Factors Affecting Emotional State and Well-being}

\textbf{Description:}  

The patient shares personal experiences, contextual factors, or circumstances that influence their health and emotional state. These include social, environmental, or personal situations beyond medical concerns that affect their emotional state.

\textbf{Examples:}
\begin{itemize}
    \item "With my father’s illness and financial stress, I’m feeling overwhelmed."
    \item "I've been under a lot of pressure at work, and now I'm having trouble sleeping."
    \item "Ever since the accident, I can't stop thinking about what happened."
    \item "I recently moved to a different state, haven't found a general practitioner, and haven't paid my high deductible for the year."
\end{itemize}

\textbf{3. Expressions of Distressing Uncertainty About Health or Treatment}

\textbf{Description:}  

Uncertainties, confusion, or mistrust about their health status, treatment, or future are leading to emotional distress. This includes questions about prognosis, treatment effectiveness, or doubt about potential outcomes that indicate or imply underlying emotional distress. The focus should not be on uncertainty alone but specifically on uncertainty that reflects or suggests emotional distress in the patient.

\textbf{Examples:}
\begin{itemize}
    \item "I'm not sure if this treatment is really working for me."
    \item "Do you think I should get a second opinion?"
    \item "Will chemo be fatal?"
    \item "Should my wife also get examined?"
    \item "Is this something that sounds like I should consider doing?"
    \item "I am wondering if I should see a doctor."
\end{itemize}

\textbf{4. Symptoms Significantly Affecting Emotional Well-being or Daily Life}

\textbf{Description:}  

The patient describes symptoms that significantly impact their emotional well-being or daily functioning, and they express or imply emotional distress because of these symptoms. The key is the emotional impact of the symptoms, not just the symptoms themselves.

\textbf{Examples:}
\begin{itemize}
    \item "My symptoms have been affecting my job for months."
    \item "I'm so tired all the time that I can't take care of my kids properly."
    \item "These migraines are making it impossible to enjoy my hobbies."
    \item "The pain is getting worse every day, and it's really wearing me down."
\end{itemize}

\subsubsection{Interpretations Not Applicable}
Interpretations are not necessary when the patient's query does not require the doctor to communicate an understanding of the patient's feelings or experiences. This occurs when:
\begin{itemize}
    \item The query is straightforward, factual, or routine.
    \item There are no expressed or implied feelings needing acknowledgment.
    \item There are no contextual factors (experiences) or underlying uncertainty concerns leading to emotional distress that require understanding.
\end{itemize}

Below are detailed categories reflecting when interpretations are not applicable:

\textbf{1. Straightforward Medical Queries Lacking Emotion, Distressing Uncertainty, and Context}

\textbf{Description:}  
The patient requests specific medical information or explanations of medical concepts without expressing emotional distress, underlying distressful uncertainty, or providing context (social, environmental, or personal situations) implying an emotional state. These queries are strictly informational and lack emotional or experiential elements requiring interpretation.

\textbf{Examples:}
\begin{itemize}
    \item "What is the use of Tylenol?"
    \item "Hello doctor, I would like to get an opinion regarding the attached chest radiograph. I wish to know if there are any abnormalities like scarring."
\end{itemize}

\textbf{2. General Health Management Requests Without Emotion, Context, and Distressing Uncertainty}

\textbf{Description:}  
The patient seeks guidance on health management, follows up on prior advice, or requests basic guidance on minor health issues without expressing emotional distress, underlying distressful uncertainty, or providing contextual factors (social, environmental, or personal situations) that imply an emotional state. Here the guidance is on what the patient should do.

\textbf{Examples:}
\begin{itemize}
    \item "I'm managing diabetes with insulin. How often should I check my blood sugar levels?"
    \item "I have intermittent knee pain from working out. How would I know if I tore cartilage?"
    \item "I had an X-ray for a fracture; should it be strapped or cast right away?"
\end{itemize}

\textbf{3. Diagnosis Requests with Neutral Symptom Descriptions Lacking Distressing Uncertainty and Context}

\textbf{Description:}  

The patient describes symptoms neutrally without expressing emotional distress or underlying distressful uncertainty. They provide necessary details without implying feelings or contextual factors (social, environmental, or personal situations) that need acknowledgment. These descriptions are straightforward and lack emotional or experiential content requiring interpretation. Here the request is about asking what the doctor thinks the issue is.

\textbf{Examples:}
\begin{itemize}
    \item "I have swelling in my ankle after a long walk. Should I be concerned?"
    \item "Hello doctor, I am suffering from pain in my mouth. It feels like sensitivity pain. I cannot say it is pain exactly; it is irritating a lot. No pain in teeth. It feels like itching in my gums (middle of the teeth). Please tell me what I can do."
\end{itemize}

\textbf{4. Hypothetical Medical Queries With No Emotions, Context, and Distressing Uncertainty}

\textbf{Description:}  

The patient inquires about hypothetical situations or general medical information without expressing or implying personal feelings or contextual factors (social, environmental, or personal situations) that need acknowledgment. 

These queries are theoretical and lack emotional or experiential aspects requiring interpretation.

\textbf{Examples:}
\begin{itemize}
    \item "If someone has XYZ symptoms, what might be the cause?"
    \item "What would happen if a person skipped their medication?"
\end{itemize}

\section{Annotation Instructions for Human Annotators}
\label{appendix:annotation_instructions}
\begin{figure*}[h]
  \centering
  \includegraphics[width=\linewidth]{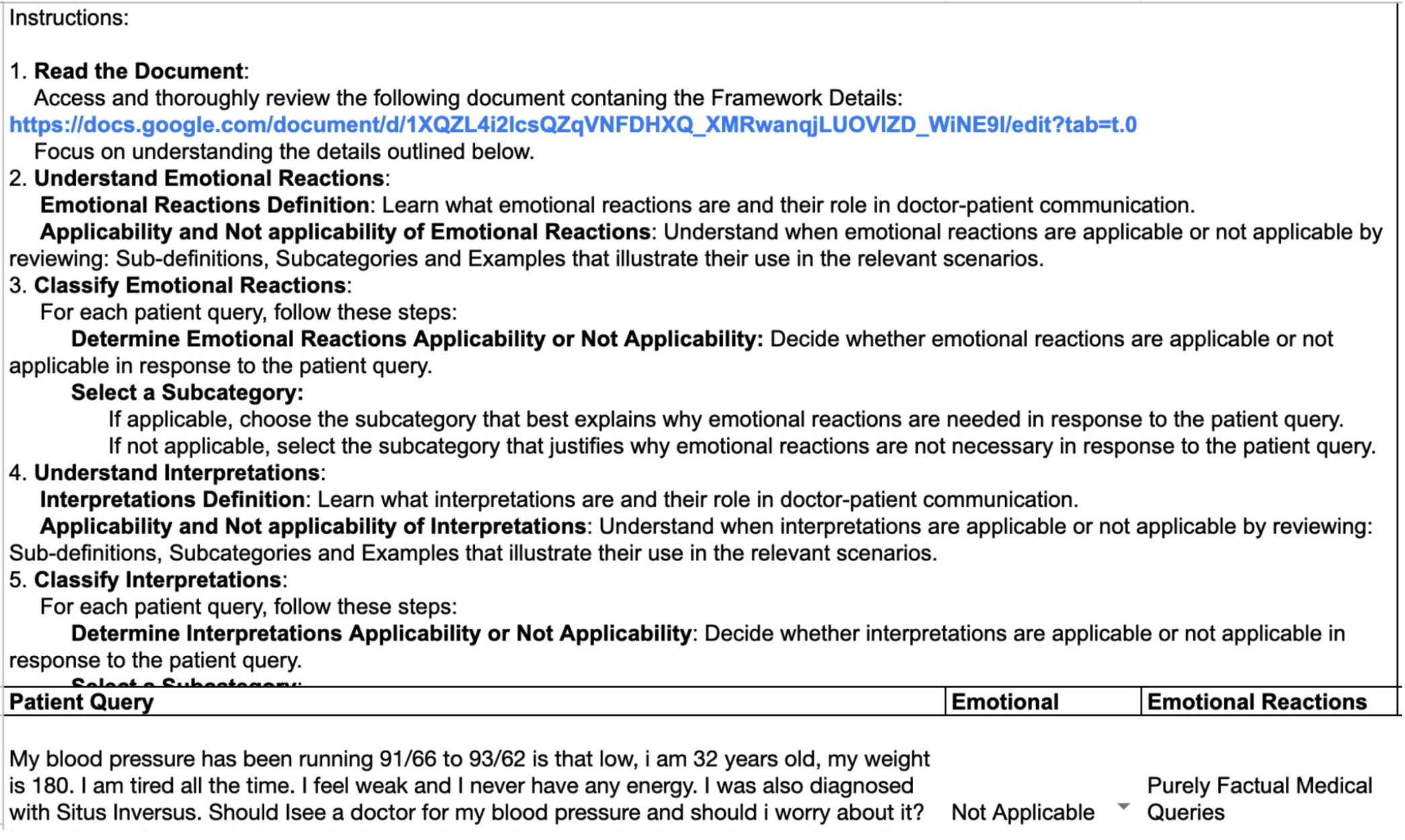}
  \caption{Screenshot of the annotation spreadsheet provided to annotators.  
  The header shows the instructions and links to the framework document.}
  \label{fig:annot_instr}
\end{figure*}

Annotators received an Excel workbook containing the patient queries and a fixed header with the instructions shown in Figure~\ref{fig:annot_instr}.  
For each \texttt{pat\_query}, they assigned \textit{Emotional Reactions} and \textit{Interpretations} labels (\texttt{Applicable} / \texttt{Not~Applicable}) and selected the justifying sub‑category, as defined in Appendix~\ref{appendix:framework_details}.  
The header also links to a Google Doc, reproduced verbatim in Appendix~\ref{appendix:framework_details}, that provides the full framework details for reference during annotation.

For accessibility and clarity, we restate the instructions here in text form.
\subsection{Instructions Given to Annotators}

\noindent\textbf{Instructions:}
\begin{enumerate}
    \item \textbf{Read the Document:} Access and thoroughly review the following document containing the Framework Details: defined in Appendix~\ref{appendix:framework_details}. \\
    Focus on understanding the details outlined below.
    
    \item \textbf{Understand Emotional Reactions:}
    \begin{itemize}
        \item \textbf{Emotional Reactions Definition:} Learn what emotional reactions are and their role in doctor-patient communication.
        \item 
        Understand when emotional reactions are applicable or not applicable by reviewing: Sub-definitions, Subcategories and Examples that illustrate their use in the relevant scenarios.
    \end{itemize}
    
    \item \textbf{Classify Emotional Reactions:} For each patient query, follow these steps:
    \begin{itemize}
        \item \textbf{Determine Emotional Reactions Applicability or Not Applicability:} Decide whether emotional reactions are applicable or not applicable in response to the patient query.
        \item \textbf{Select a Subcategory:}
        \begin{itemize}
            \item If applicable, choose the subcategory that best explains why emotional reactions are needed in response to the patient query.
            \item If not applicable, select the subcategory that justifies why emotional reactions are not necessary in response to the patient query.
        \end{itemize}
    \end{itemize}
    
    \item \textbf{Understand Interpretations:}
    \begin{itemize}
        \item \textbf{Interpretations Definition:} Learn what interpretations are and their role in doctor-patient communication.
        \item 
        Understand when interpretations are applicable or not applicable by reviewing: Sub-definitions, Subcategories and Examples that illustrate their use in the relevant scenarios.
    \end{itemize}
    
    \item \textbf{Classify Interpretations:} For each patient query, follow these steps:
    \begin{itemize}
        \item \textbf{Determine Interpretations Applicability or Not Applicability:} Decide whether interpretations are applicable or not applicable in response to the patient query.
        \item \textbf{Select a Subcategory:}
        \begin{itemize}
            \item If applicable, choose the subcategory that best explains why interpretations are needed in response to the patient query.
            \item If not applicable, select the subcategory that justifies why interpretations are not necessary in response to the patient query.
        \end{itemize}
    \end{itemize}
\end{enumerate}

\subsection{Additional Verbal Clarifications}

During training sessions, annotators received the following clarifications:

\begin{itemize}
    \item If they could not understand whether the symptoms or medical issue is \textbf{severe}, they were allowed to briefly search online (e.g., Google) to check whether the condition is typically serious.
    \item If they were unsure whether the query was emotionally significant for the patient, they were encouraged to \textbf{go with what they felt} and believe their own judgment.
    \item For each dimension (EA and IA), annotators were instructed to:
    \begin{enumerate}
        \item Read the patient query.
        \item While annotating a dimension, first read the \emph{Applicable} definition. If they believe it fits, go through the Applicable subcategories one by one and tag at least the one they think fits best.
        \item If the Applicable definition does not feel like it fits, they should still briefly review the Applicable subcategories to verify this.
        \item Then move to the \emph{Not Applicable} definition and repeat the same process with the Not Applicable subcategories.
    \end{enumerate}
\end{itemize}

\subsection{Boundary Cases: Subjectivity and Lack of Medical Expertise}

Empathy applicability judgments are inherently subjective, and some patient queries lie at the boundary between emotional and informational intent. Such disagreements often reflect \textbf{legitimate interpretive variability} rather than simple annotation error. These challenges are explored in detail in Section~\ref{challenges} (Systematic Challenges in Operationalizing Anticipatory Empathy); here, we briefly revisit representative cases to make these boundary conditions more transparent.


To make these cases transparent, we provide \texttt{misalignment\_analysis.csv} at \url{https://github.com/shanmrandhawa/Empathy-Applicability-Framework}, listing detailed patient queries, mis-aligned labels, and qualitative interpretations. The queries highlighted as exhibiting \emph{Divergent Interpretation} correspond to \emph{Reasonable Disagreement} around both symptom severity and whether emotional content is present.

\section{Illustrative Scenarios for EAF Operationalization}
\label{sec:appendix-scenarios}

See Table~\ref{tab:empathy_scenarios} for illustrative scenarios demonstrating the operationalization of the EAF.

\begin{table*}
\centering
\resizebox{\textwidth}{!}{%
\begin{tabular}{|p{2.5cm}|p{2.5cm}|p{4.5cm}|p{2.2cm}|p{4.5cm}|p{5cm}|}
\hline
\textbf{Empathy Dimension} & \textbf{Scenario Type} & \textbf{Scenario} & \textbf{Applicability} & \textbf{Explanation} & \textbf{Responses with Applicability Signals Realized} \\ \hline

Emotional Reaction & Explicit Need & 
\textit{``Hello doctor, I am having constant eye floaters, low back and hip pain, and also my rib cage hurts. I feel depressed and anxious like never before. I cannot sleep at night. An MRI of my brain shows a tiny flare, but radiologists say it's nothing to worry about. What should I do?''} & 
Applicable & 
The patient explicitly expresses intense negative emotions, such as feeling depressed and anxious and states an inability to sleep. An emotional reaction from the doctor is necessary to provide support and reassurance.&
\textit{``I'm  sorry to hear that you're going through a difficult time. Dealing with multiple symptoms while feeling depressed and anxious must be overwhelming.} I'd recommend scheduling a follow-up with your primary care physician to address the pain comprehensively, and discussing your sleep and mood concerns-a referral for mental health support may also be helpful.'' \\ \hline

Emotional Reaction & Implicit Need & 
\textit{``Hello doctor, my son has been experiencing frequent headaches over the past week. We've tried over-the-counter medications, but there's no improvement. What should we do?''} & 
Applicable & 
Emotional reactions are applicable here because, as Richert et al. \citep{richert2018being} find, parents of children with health (drug) issues often experience significant distress and negative mental health effects. The mother may be experiencing worry and anxiety about her child's well-being, even if she doesn't explicitly express it. &
\textit{``I can imagine how worrying it must be when your child's headaches persist despite medication.} I'd suggest keeping a headache diary noting triggers, and scheduling a pediatric appointment for further evaluation, especially given the persistence over a week.'' \\ \hline

Emotional Reaction & Not Needed & 
\textit{``Hello doctor, I was suffering from an infection in my tonsil for the past four days. I went to an ENT specialist who prescribed antibiotics. Now my tonsil pain has subsided, but I still feel something stuck on the left side of my throat where the pain was. I have no problem swallowing. Kindly advise me on what to do next.''} & 
Not Applicable & 
The patient provides a neutral description of symptoms without expressing emotional concern or distress. The primary need is factual medical advice. An emotional reaction from the doctor is not necessary in this case. &
``The residual sensation you're describing is common after a tonsil infection and usually resolves as the inflammation fully subsides. Continue your antibiotics as prescribed, stay hydrated, and if the sensation persists beyond a week, follow up with your ENT specialist.'' \\ \hline

Interpretation & Explicit Need & 
\textit{``Hello doctor, I am feeling extremely anxious about my upcoming surgery. I can't stop worrying about the possible complications.''} & 
Applicable & 
The patient explicitly expresses feelings of anxiety and worry. The doctor should communicate an understanding of these feelings, acknowledging the patient's emotional state and providing appropriate support. &
\textit{``It sounds like the uncertainty about what could happen during surgery is weighing heavily on you, and that kind of worry before a procedure is very natural.} Let's go over the specific risks together so you have a clear picture; often, understanding the actual likelihood of complications can help ease some of that anxiety.'' \\ \hline

Interpretation & Implicit Need & 
\textit{``Hello doctor, I've been taking the medication as prescribed, but I'm not seeing any improvement. Is there something I'm doing wrong?''} & 
Applicable & 
The patient implies feelings of frustration and possibly self-blame. The doctor should interpret and acknowledge these underlying feelings, demonstrating understanding and support. &
\textit{``It sounds like you may be feeling uncertain that things haven't improved, and perhaps wondering if you're somehow at fault; I want you to know that lack of response isn't necessarily something you caused.} Some medications take longer to show effects, and there may be alternative treatments we can explore. Let's review your current regimen together.'' \\ \hline

Interpretation & Not Needed & 
\textit{``I was playing with my sister's boyfriend's brother and I swung to hit him like I said we were playing around and my wrist hit his elbow really hard when it happened my hand got really numb and my vein was hurting really bad and it's 6 hours later and my vein still hurts what should I do''} & 
Not Applicable & 
The query is a straightforward request for diagnosis with neutral symptom descriptions. It does not express emotions or distressing contextual factors that require acknowledgment. The doctor's response should focus solely on providing a factual diagnosis. &
``Based on your description, you may have bruised the soft tissue or a superficial vein in your wrist. Apply ice for 15--20 minutes at a time, keep the wrist elevated, and take an over-the-counter anti-inflammatory if needed. If swelling, discoloration, or numbness worsens, seek in-person medical evaluation to rule out a fracture.'' \\ \hline

\end{tabular}%
}
\caption{Empathy Dimensions, Scenarios, Applicability, and Explanations with Example Realized Responses}
\label{tab:empathy_scenarios}
\end{table*}

\section{Human-GPT Agreement Analysis}
\label{appendix:human_gpt_agreement}

Table~\ref{tab:kappa_human_gpt_scores} presents pairwise agreement between GPT and each human annotator. “Agreed” and “Disagreed” columns denote the number of queries where both annotators assigned the same or different labels of Applicable or Not Applicable, respectively.

\begin{table*}[h]
    \centering
    \caption{Cohen’s $\kappa$ agreement scores and confusion matrix counts between GPT-4o and each human annotator for Emotional Reactions Applicability (EA) and Interpretations Applicability (IA).}
    \label{tab:kappa_human_gpt_scores}
    \resizebox{\textwidth}{!}{%
    \begin{tabular}{lcccccccc}
        \hline
        Annotator 1 & Annotator 2 & Kappa EA & Kappa IA & Agreed EA & Disagreed EA & Agreed IA & Disagreed IA \\
        \hline
        HA1 & GPT & 0.4402 & 0.5306 & 917 & 379 & 988 & 308 \\
        HA2 & GPT & 0.4096 & 0.3612 & 940 & 356 & 890 & 406 \\
        \hline
    \end{tabular}%
    }
\end{table*}

\section{Model Architecture Details} \label{appendix:model_architecture}

\begin{figure*}[h] \centering \includegraphics[width=1\textwidth]{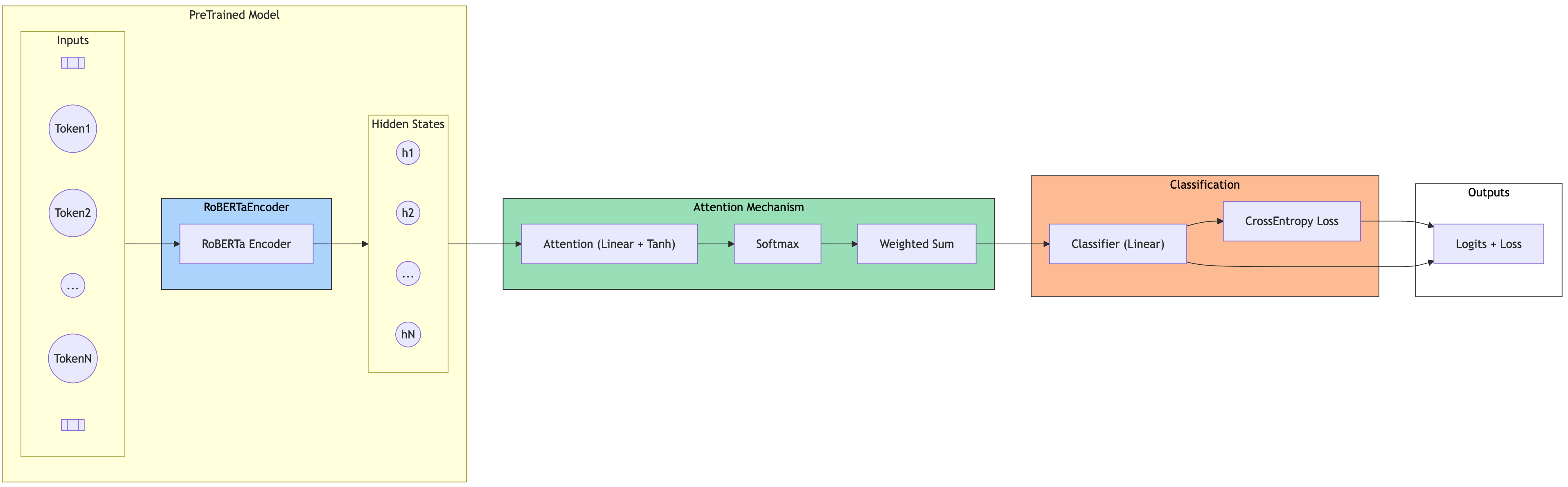} \caption{Empathy Dimension Applicability Model Architecture} \label{fig:single_dimension_model} \end{figure*}

Each empathy dimension, Emotional Reactions (EA) and Interpretations (IA), is modeled independently. We fine-tune a pretrained RoBERTa-based model \citep{liu2019roberta} separately for each dimension, while maintaining the same overall architecture. “Independently” means each classifier learns to predict the applicability of one dimension without sharing parameters or optimization across tasks. For fine-tuning, we incorporate an attention mechanism based on a feed-forward network. The model architecture is illustrated in Figure~\ref{fig:single_dimension_model}.

The model follows an attention-based pooling approach built on top of a pretrained RoBERTa encoder. The encoder converts patient queries into contextualized token embeddings, capturing the meaning of each word based on its surrounding context. When a sentence is processed by RoBERTa, it generates a hidden representation for each token, reflecting its contextual meaning. Unlike traditional methods that rely solely on the [CLS] token or an average of all embeddings, this model applies a learned attention mechanism to identify the most relevant tokens for classification.

Specifically, the model uses a feed-forward neural network to compute attention scores for each token. A linear transformation first maps each token embedding to a scalar score, which then passes through a Tanh activation to constrain values between \([-1, 1]\) and avoid extremes. Since not all tokens contribute equally to classification, the model converts these raw scores into attention weights using a softmax function across the sequence. This normalization ensures that important words receive higher weights, while less relevant words are assigned lower importance.

After computing attention weights, the model performs a weighted sum of token embeddings. Tokens with higher attention scores contribute more significantly to the final pooled representation, highlighting the most relevant parts of the query. This pooled vector is then passed through a classification-linear layer, which outputs logits representing the likelihood of belonging to either the "Not Applicable" or "Applicable" class. During training, the model optimizes both the attention mechanism and the classification layer via cross-entropy loss, thereby improving accuracy in empathy classification.

Training separate models for EA and IA avoids crosstalk between tasks. Each classifier learns dimension-specific patterns from the data, resulting in a simple and modular approach that enables focused analysis of empathy applicability in patient queries.

\section{Prompt Design for LLM Annotations}
\label{app:prompt-design}
For more detail on the prompt design used for LLM (GPT-4o, o1) based annotations, we provide here the exact prompts used in our experiments.

We used two styles of prompts:

\begin{itemize}
    \item \textbf{With-framework (contrastive) prompts}: the LLM received the full Empathy Applicability Framework for a given dimension (Emotional Reactions or Interpretations), including both Applicable and Not Applicable subcategories with examples for each. This creates a contrastive in-context signal: the model must decide between multiple subcategories across both classes.
    \item \textbf{Without-framework prompts}: the LLM only received a short task definition (dimension definition + binary Applicability decision), without any subcategories or examples. This approximates a generic zero-shot setup without our framework.
\end{itemize}

\subsection{Emotional Reactions: With-Framework Contrastive Prompt}

The core schema passed to the LLM for Emotional Reactions with the full framework was:

\begin{lstlisting}
ANNOTATION_SCHEMA = {
   "instruction": "Annotate emotional reactions in general health queries based on the following schema. For each query, return the matching subcategories. Think logically and ensure to revisit your annotation for each query",
   "definitions": {
       "Emotional Reactions": {
           "description": "Expressions of warmth, compassion, concern, or similar feelings conveyed by a doctor in response to a patient's query.",
           "categories": [
               {
                   "name": "Purely Factual Medical Queries",
                   "description": "The patient requests specific medical information, including explanations of medical concepts, without emotional distress or underlying distressing uncertainty.",
                   "examples": ["What is the use of Tylenol?", "Is it possible to outgrow a seafood allergy?"],
                   "class" : "Emotional Reactions Not Applicable"
               },
               {
                   "name": "General Health Management Without Emotional Involvement",
                   "description": "The patient seeks guidance on health management, follows up on prior advice, or requests basic guidance on minor health issues, without expressing emotional distress or underlying distressing uncertainty. Here the guidance is on what the patient should do.",
                   "examples": ["I'm managing diabetes with insulin. How often should I check my blood sugar levels?", "I have swelling in my ankle after a long walk. Should I be concerned?"],
                   "class" : "Emotional Reactions Not Applicable"
               },
               {
                   "name": "Diagnosis Requests with Neutral Symptom Descriptions",
                   "description": "The patient describes symptoms neutrally without expressing emotional distress or underlying distressing uncertainty. Here the request is about asking what the doctor thinks the issue is.",
                   "examples": ["I have intermittent knee pain from working out. How would I know if I tore cartilage?", "Hello. I am having pain in my jaw area, immediately in front of my left ear."],
                   "class" : "Emotional Reactions Not Applicable"
               },
               {
                   "name": "Hypothetical Medical Queries Without Emotional Concern",
                   "description": "The patient inquires about hypothetical situations without emotional involvement.",
                   "examples": ["If someone has XYZ symptoms, what might be the cause?", "What would happen if a person skipped their medication?"],
                   "class" : "Emotional Reactions Not Applicable"
               },
               {
                   "name": "Seriousness of Symptoms",
                   "description": "The patient describes symptoms that suggest a life-threatening or chronic health condition significantly impacting long-term health or quality of life. This includes diseases like cancer, heart disease, mental health issues, or chronic conditions leading to disability.",
                   "examples": ["My father has been having severe chest pains and shortness of breath. Could it be a heart attack?", "I've been experiencing numbness and weakness in my limbs for months."],
                   "class" : "Emotional Reactions Applicable"
               },
               {
                   "name": "Severe Negative Emotion Expressed",
                   "description": "The patient explicitly states intense emotions such as fear, frustration, or anger regarding their health.",
                   "examples": ["I feel depressed and anxious like never before. I cannot sleep at night.", "I'm terrified about my recent diagnosis of cancer."],
                   "class" : "Emotional Reactions Applicable"
               },
               {
                   "name": "Underlying Negative Emotional State Inferred",
                   "description": "The patient implies emotional distress that isn't explicitly stated but can be inferred from their tone or descriptions, such as subtle signs of emotional worry, frustration, or distress about delays or underlying distressing uncertainties. Focus on emotional worry, not the medical concern.",
                   "examples": ["I am starting to get a little alarmed by this spotting after ovulation. Is this cause for concern?", "I need to be a bit more at ease after what I read about diabetic enteropathy."],
                   "class" : "Emotional Reactions Applicable"
               },
               {
                   "name": "Concern Severity for Close Relations",
                   "description": "The patient is asking on behalf of someone with whom they share a close, protective relationship, implying heightened emotional concern.",
                   "examples": ["Hello, I am the mother of a five-year-old. He has a small lump that hasn't gone away.", "My son recently started daycare and has gotten sick. His fever was 102.9. Should I take him to the hospital?"],
                   "class" : "Emotional Reactions Applicable"
               }
           ]
       }
   },
   "output_format": "json",
   "example_query": {
       "query": "I'm scared and plan on taking my son to the doctor. Should I be overly worried?",
       "annotations": [
         {
           "subcategories": [
             {"name": "Severe Negative Emotion Expressed"},
             {"name": "Concern Severity for Close Relations"}
           ],
           "class": "Emotional Reactions Applicable",
           "reason": "The patient explicitly expresses fear regarding their son's health and shows heightened emotional concern for a close relation."
         }
       ]
   }
}
\end{lstlisting}

\subsection{Interpretations: With-Framework Contrastive Prompt}

The corresponding schema for Interpretations (IA) with the full framework was:

\begin{lstlisting}
ANNOTATION_SCHEMA = {
   "instruction": "Annotate interpretations in general health queries based on the following schema. For each query, return the matching subcategories. Think logically and ensure to revisit your annotation for each query",
   "definitions": {
       "Interpretations": {
           "description": "Interpretations refer to the communication of an understanding of the patient's feelings (explicit or implied) and/or experiences (contextual factors) inferred from their query. It's about recognizing and articulating what the patient is feeling and why, based on their situation, concerns, and history.",
           "categories": [
               {
                   "name": "Expression of Feelings (Explicit or Implicit)",
                   "description": "The patient expresses emotions directly or implies them through language or tone. This includes feelings such as fear, anxiety, frustration, sadness, or hopelessness.",
                   "examples": [
                       "I'm really scared about these chest pains.",
                       "I'm frustrated because my symptoms aren't improving.",
                       "I guess I have to accept this is how things will be now.",
                       "Nothing seems to be helping.",
                       "I don't know what to do anymore."
                   ],
                   "class": "Interpretations Applicable"
               },
               {
                   "name": "Sharing of Experiences or Contextual Factors Affecting Emotional State and Well being",
                   "description": "The patient shares personal experiences, contextual factors, or circumstances that influence their health and emotional state. These include social, environmental, or personal situations, beyond medical concerns, that affect their emotional state.",
                   "examples": [
                       "With my father's illness and financial stress, I'm feeling overwhelmed.",
                       "I've been under a lot of pressure at work, and now I'm having trouble sleeping.",
                       "Ever since the accident, I can't stop thinking about what happened.",
                       "I recently moved to a different state, haven't found a general practitioner, and haven't paid my high deductible for the year."
                   ],
                   "class": "Interpretations Applicable"
               },
               {
                   "name": "Expressions of Distressing Uncertainty About Health or Treatment",
                   "description": "Uncertainties, confusion, or mistrust expressed by a patient about their health status, treatment, or future that lead to significant emotional distress. This includes statements involving questions about prognosis, treatment effectiveness, or doubt about potential outcomes, specifically when accompanied by explicit or implicit signs of emotional distress.",
                   "examples": [
                       "I'm not sure if this treatment is really working for me, and it's making me anxious.",
                       "Is this something that sounds like I should consider doing? I'm so confused about what's right.",
                       "I feel lost. Should my wife also get examined?",
                       "Do you think there's any hope for me after trying this?"
                   ],
                   "class": "Interpretations Applicable"
               },
               {
                   "name": "Symptoms Significantly Affecting Emotional Well-being or Daily Life",
                   "description": "The patient describes symptoms that significantly impact their emotional well-being or daily functioning, and they express or imply emotional distress because of these symptoms. The key is the emotional impact of the symptoms, not just the symptoms themselves.",
                   "examples": [
                       "These migraines are making it impossible to enjoy my hobbies.",
                       "I'm so tired all the time that I can't take care of my kids properly.",
                       "My symptoms have been affecting my job for months.",
                       "The pain is getting worse every day, and it's really wearing me down."
                   ],
                   "class": "Interpretations Applicable"
               },
               {
                   "name": "Straightforward Medical Queries Lacking Emotion, Distressing Uncertainty, and Context",
                   "description": "The patient requests specific medical information or explanations of medical concepts without expressing emotional distress, underlying distressful uncertainty or providing context (social, environmental, or personal situations) implying an emotional state. These queries are strictly informational and lack emotional or experiential elements requiring interpretation.",
                   "examples": [
                       "What is the use of Tylenol?",
                       "Hello doctor, I would like to get an opinion regarding the attached chest radiograph. I wish to know if there are any abnormalities like scarring."
                   ],
                   "class": "Interpretations Not Applicable"
               },
               {
                   "name": "General health management requests Without Emotion, Context, and Distressing Uncertainty",
                   "description": "The patient seeks guidance on health management, follows up on prior advice, or requests basic guidance on minor health issues without expressing emotional distress, underlying distressful uncertainty, or providing contextual factors that imply an emotional state. Additionally, they may include personal medical context, such as test results, medications taken, and previous medical consultations. Here the guidance is on what the patient should do.",
                   "examples": [
                       "I have intermittent knee pain from working out. How would I know if I tore cartilage?",
                       "I had an X-ray for a fracture; should it be strapped or cast right away?",
                       "Hi, my husband is 39, and his SGPT and SGOT levels in a recent test were 101 and 98 respectively. His triglycerides are 280, which is high. His height is 168 cm and weight is 79 kg. What does a rise in these values indicate? What precautions should he take?"
                   ],
                   "class": "Interpretations Not Applicable"
               },
               {
                   "name": "Diagnosis Requests with Neutral Symptom Descriptions Lacking Distressing Uncertainty and Context",
                   "description": "The patient describes symptoms neutrally without expressing emotional distress or underlying distressful uncertainty. They provide necessary details without implying feelings or contextual factors. These descriptions are straightforward and lack emotional or experiential content requiring interpretation. Here the request is about asking what the doctor thinks the issue is.",
                   "examples": [
                       "I have swelling in my ankle after a long walk. Should I be concerned?",
                       "Hello doctor, I am suffering from pain in my mouth. It feels like sensitivity pain. I cannot say it is pain exactly; it is irritating a lot. No pain in teeth. It feels like itching in my gums (middle of the teeth). Please tell me what I can do."
                   ],
                   "class": "Interpretations Not Applicable"
               },
               {
                   "name": "Hypothetical Medical Queries with no Emotions, Context, and Distressing Uncertainty",
                   "description": "The patient inquires about hypothetical situations or general medical information without expressing or implying personal feelings or contextual factors that need acknowledgment. These queries are theoretical and lack emotional or experiential aspects requiring interpretation.",
                   "examples": [
                       "If someone has XYZ symptoms, what might be the cause?",
                       "What would happen if a person skipped their medication?"
                   ],
                   "class": "Interpretations Not Applicable"
               }
           ]
       }
   },
   "output_format": "json",
   "example_query": {
       "query": "I'm not sure if this treatment is really working for me.",
       "annotations": [
           {
               "subcategories": [
                   {
                       "name": "Expressions of Distressing Uncertainty About Health or Treatment"
                   }
               ],
               "class": "Interpretations Applicable",
               "reason": "The patient explicitly expresses doubt about the effectiveness of the treatment, which requires interpretation."
           }
       ]
   }
}
\end{lstlisting}

\subsection{Prompts Without the Framework (Definition-Only)}

For the \emph{without-framework} condition, the LLM received only a short task description and the names of the Applicability labels. No subcategories or examples were provided.

\subsubsection{Emotional Reactions (without framework).}

\begin{lstlisting}
ANNOTATION_SCHEMA = {
   "instruction": "Read the patient query and decide whether emotional reactions are necessary in the response. "
                  "Emotional reactions refer to the expressions of warmth, compassion, concern, or similar feelings "
                  "conveyed by a doctor in response to a patient's query. "
                  "If emotional reactions are necessary, mark it as 'Emotional Reactions Applicable'. "
                  "If not, mark it as 'Emotional Reactions Not Applicable'. Think carefully and be consistent.",
   "output_format": "json"
}
\end{lstlisting}

\subsubsection{Interpretations (without framework).}

\begin{lstlisting}
ANNOTATION_SCHEMA = {
   "instruction": "Read the patient query and decide whether interpretations are necessary in the response. "
                  "Interpretations refer to the communication of an understanding of the patient's feelings "
                  "(explicit or implied) and/or experiences (contextual factors) inferred from their query. "
                  "If interpretations are necessary, mark it as 'Interpretations Applicable'. "
                  "If not, mark it as 'Interpretations Not Applicable'. Think carefully and be consistent.",
   "output_format": "json"
}
\end{lstlisting}

Together, these listings document the exact prompts used in both the with-framework (contrastive) and without-framework settings for LLM annotations.


\section{Dataset Analyses}
\label{app:dataset-analyses}

\begin{figure*}[t]
    \centering
    \includegraphics[width=\textwidth]{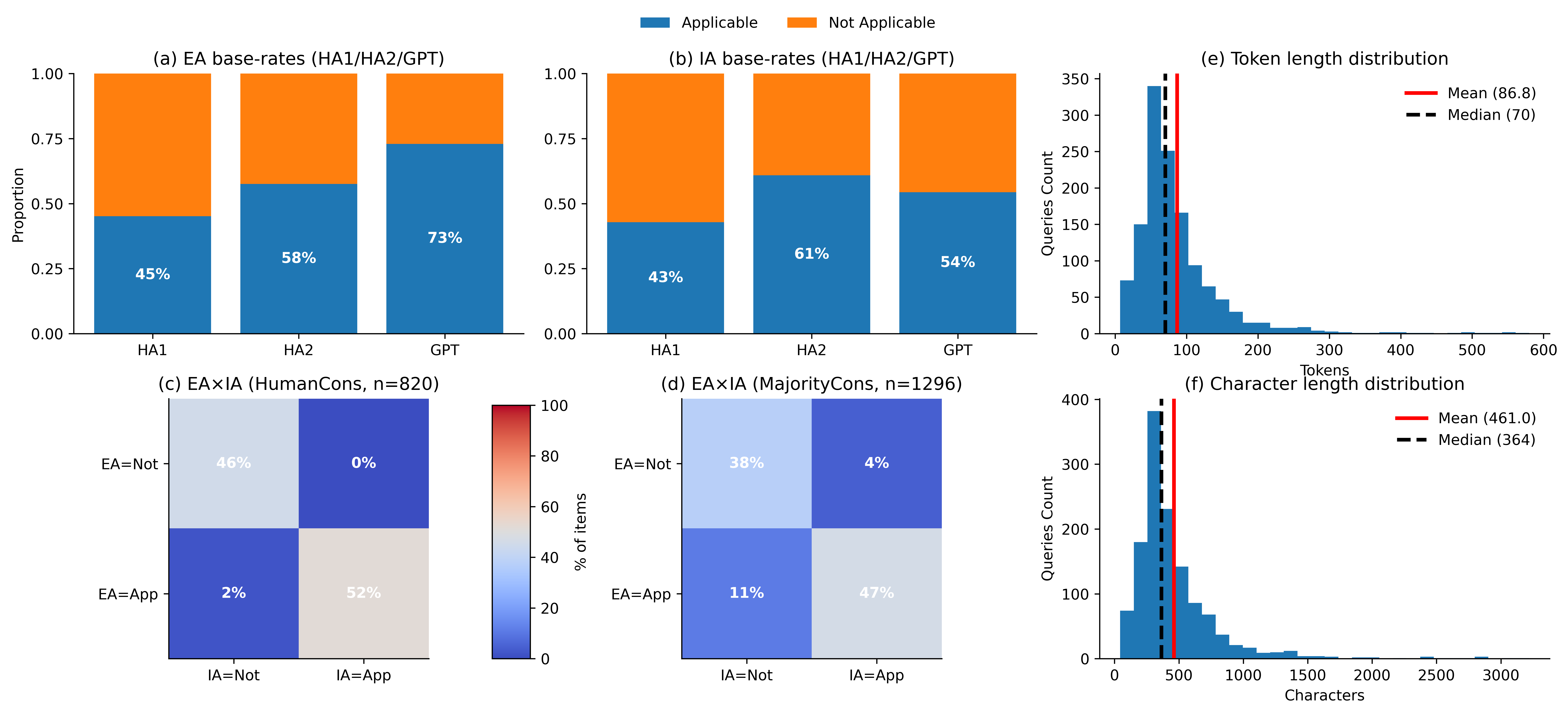}
    \caption{\textbf{Dataset overview panel: base rates, EA--IA coupling, and query length distributions.}
    Panels (a--b) report the binary label base rates for \emph{Emotional Reactions (EA)} and \emph{Interpretations (IA)} from Human Annotator 1 (HA1), Human Annotator 2 (HA2), and GPT. Bars are shown as stacked proportions of \textit{Applicable} vs.\ \textit{Not Applicable}.
    Panels (c--d) show EA$\times$IA co-occurrence as 2$\times$2 heatmaps for (c) \textit{Human consensus} (only items where HA1 and HA2 agree on both EA and IA) and (d) \textit{Majority consensus} (majority vote over HA1, HA2, and GPT), with each cell annotated by the percentage of items in that consensus subset; 
    Panels (e--f) summarize query length: (e) token-count histogram (simple word tokenization) and (f) character-count histogram; vertical reference lines mark the mean (solid red) and median (dashed black).
    Together, the figure summarizes label prevalence, the empirical coupling between EA and IA decisions, and the distribution of textual input lengths in patient queries.}
    \label{fig:dataset_overview_panel}
\end{figure*}

To characterize the released benchmark beyond agreement and modeling results, Figure~\ref{fig:dataset_overview_panel} summarizes label base rates (H1/H2/GPT), EA--IA co-applicability patterns, and the heavy-tailed distribution of query lengths, providing a high-level view of dataset variability. In the following sections, we present additional analyses of annotation consistency, framework coherence, and stability across the labeling process. Specifically, we analyze: (i) subcategory usage and co-occurrence patterns across human annotators and GPT rationales; (ii) length effects on applicability with confidence intervals; (iii) run-order drift checks to assess stability over the labeling sequence; and (iv) match vs.\ miss portions of subcategory rationales when humans and GPT agree on the overall applicability label, to assess alignment in rationales.


\begin{table*}[t]
\centering
\small
\setlength{\tabcolsep}{5pt}
\renewcommand{\arraystretch}{1.15}
\begin{tabular}{p{4.5cm} p{2.45cm} c r r r}
\toprule
\textbf{EA subcategory} & \textbf{Shortname} & \textbf{Class} & \textbf{HA1} & \textbf{HA2} & \textbf{GPT} \\
\midrule
Underlying Negative Emotional State Inferred & inferred negative emotion & App & 121 (9.3) & 197 (15.2) & 770 (59.4) \\
Concern Severity For Close Relations & concern for relations & App & 246 (19.0) & 267 (20.6) & 277 (21.4) \\
Severe Negative Emotion Expressed & severe negative emotion & App & 209 (16.1) & 132 (10.2) & 177 (13.7) \\
Seriousness Of Symptoms & symptom seriousness & App & 9 (0.7) & 150 (11.6) & 357 (27.5) \\
\addlinespace
Diagnosis Requests With Neutral Symptom Descriptions & neutral diagnosis request & Not & 262 (20.2) & 165 (12.7) & 226 (17.4) \\
Purely Factual Medical Queries & factual medical query & Not & 412 (31.8) & 38 (2.9) & 108 (8.3) \\
General Health Management Without Emotional Involvement & general health management & Not & 37 (2.9) & 252 (19.4) & 156 (12.0) \\
Hypothetical Medical Queries Without Emotional Concern & hypothetical medical query & Not & 0 (0.0) & 95 (7.3) & 29 (2.2) \\
\bottomrule
\end{tabular}
\caption{EA subcategory prevalence with applicability class (App vs.\ Not). Values are count (percent of $N=1296$). Humans assign one subcategory per query (counts sum to $N$). GPT may assign multiple subcategories per query; thus GPT counts can exceed $N$.}
\label{tab:ea-prevalence-app}
\end{table*}

\begin{table*}[t]
\centering
\small
\setlength{\tabcolsep}{5pt}
\renewcommand{\arraystretch}{1.15}
\begin{tabular}{p{4.8cm} p{2.45cm} c r r r}
\toprule
\textbf{IA subcategory} & \textbf{Shortname} & \textbf{Class} & \textbf{HA1} & \textbf{HA2} & \textbf{GPT} \\
\midrule
Expressions Of Distressing Uncertainty About Health Or Treatment & distressing uncertainty & App & 222 (17.1) & 529 (40.8) & 556 (42.9) \\
Expression Of Feelings (Explicit Or Implicit) & feelings expression & App & 274 (21.1) & 176 (13.6) & 423 (32.6) \\
Symptoms Significantly Affecting Emotional Well-Being Or Daily Life & impact on daily life & App & 44 (3.4) & 36 (2.8) & 382 (29.5) \\
Sharing Experiences Or Contextual Factors Affecting Emotional State And Well Being & sharing affecting context & App & 15 (1.2) & 48 (3.7) & 308 (23.8) \\
\addlinespace
Straightforward Medical Queries Lacking Emotion, Distressing Uncertainty, And Context & straightforward medical query & Not & 431 (33.3) & 41 (3.2) & 285 (22.0) \\
Diagnosis Requests With Neutral Symptom Descriptions Lacking Distressing Uncertainty And Context & neutral diagnosis request & Not & 269 (20.8) & 118 (9.1) & 339 (26.2) \\
General Health Management Requests Without Emotion, Context, And Distressing Uncertainty & general health management & Not & 41 (3.2) & 264 (20.4) & 242 (18.7) \\
Hypothetical Medical Queries With No Emotions, Context, And Distressing Uncertainty & hypothetical medical query & Not & 0 (0.0) & 84 (6.5) & 37 (2.9) \\
\bottomrule
\end{tabular}
\caption{IA subcategory prevalence with applicability class (App vs.\ Not). Values are count (percent of $N=1296$). Humans assign one subcategory per query (counts sum to $N$). GPT may assign multiple subcategories per query; thus GPT counts can exceed $N$.}
\label{tab:ia-prevalence-app}
\end{table*}

\begin{figure*}[t]
\centering
\begin{subfigure}[t]{0.32\textwidth}
  \centering
  \includegraphics[width=\linewidth]{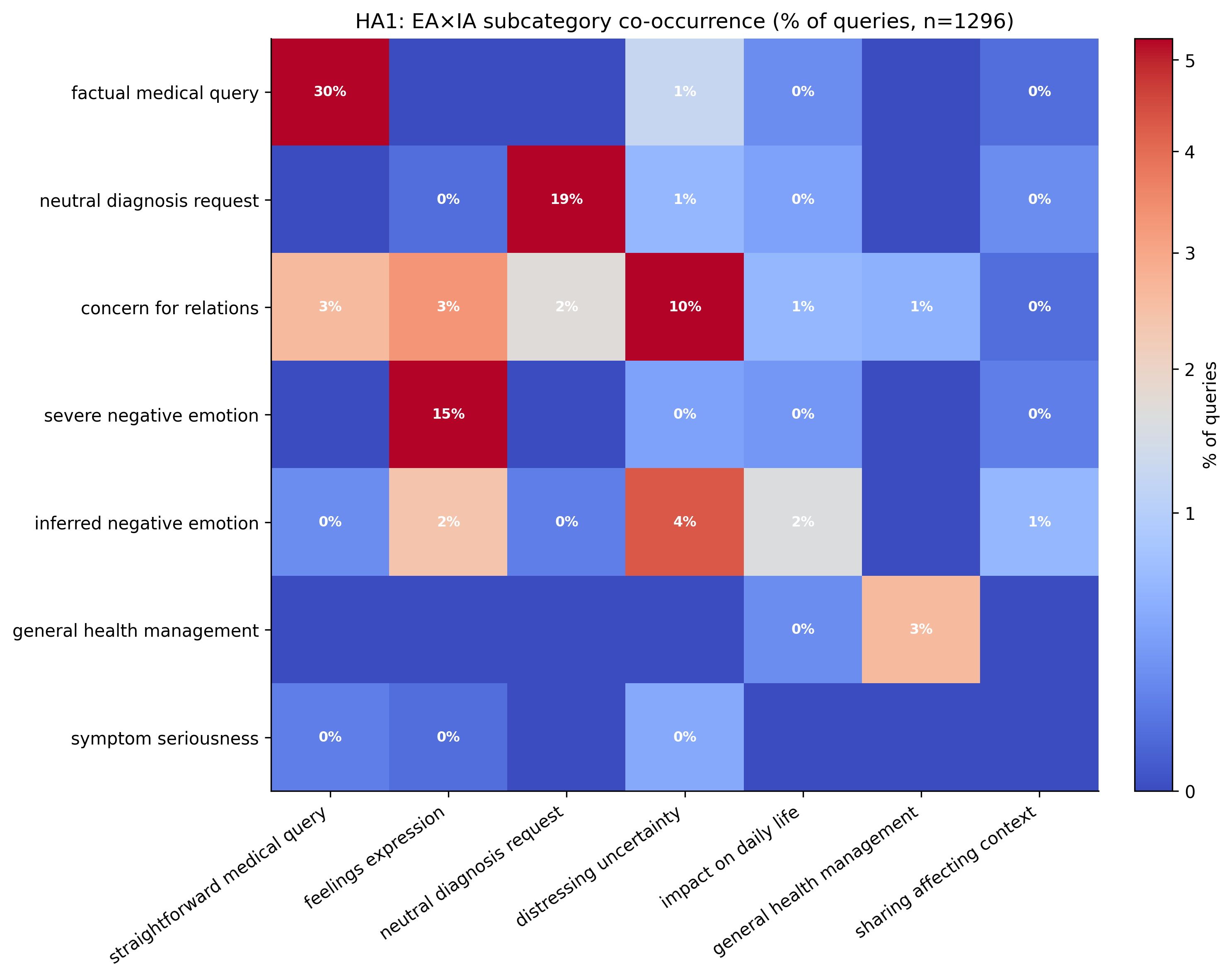}
  \caption{HA1: EA$\times$IA co-occurrence.}
\end{subfigure}\hfill
\begin{subfigure}[t]{0.32\textwidth}
  \centering
  \includegraphics[width=\linewidth]{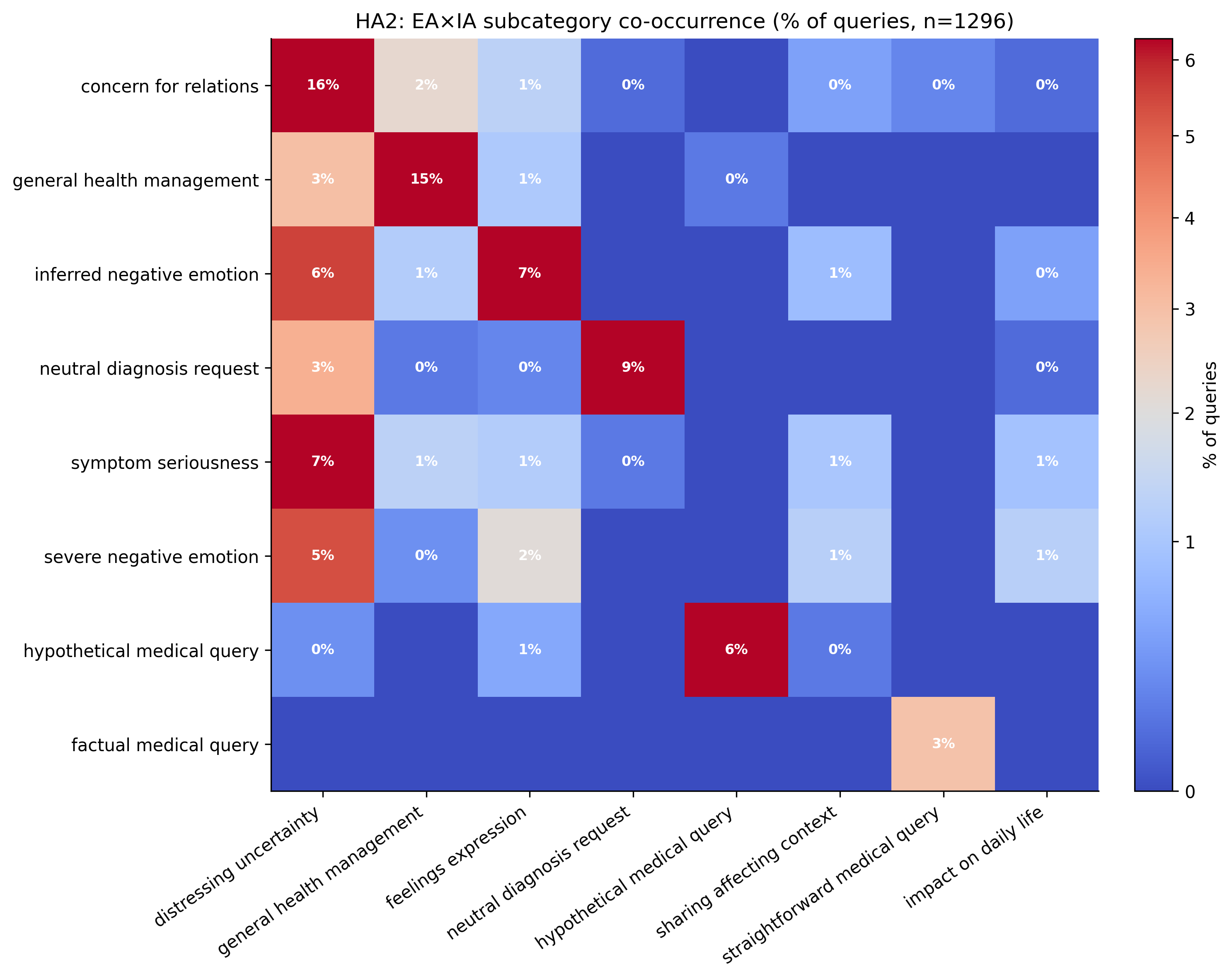}
  \caption{HA2: EA$\times$IA co-occurrence.}
\end{subfigure}\hfill
\begin{subfigure}[t]{0.32\textwidth}
  \centering
  \includegraphics[width=\linewidth]{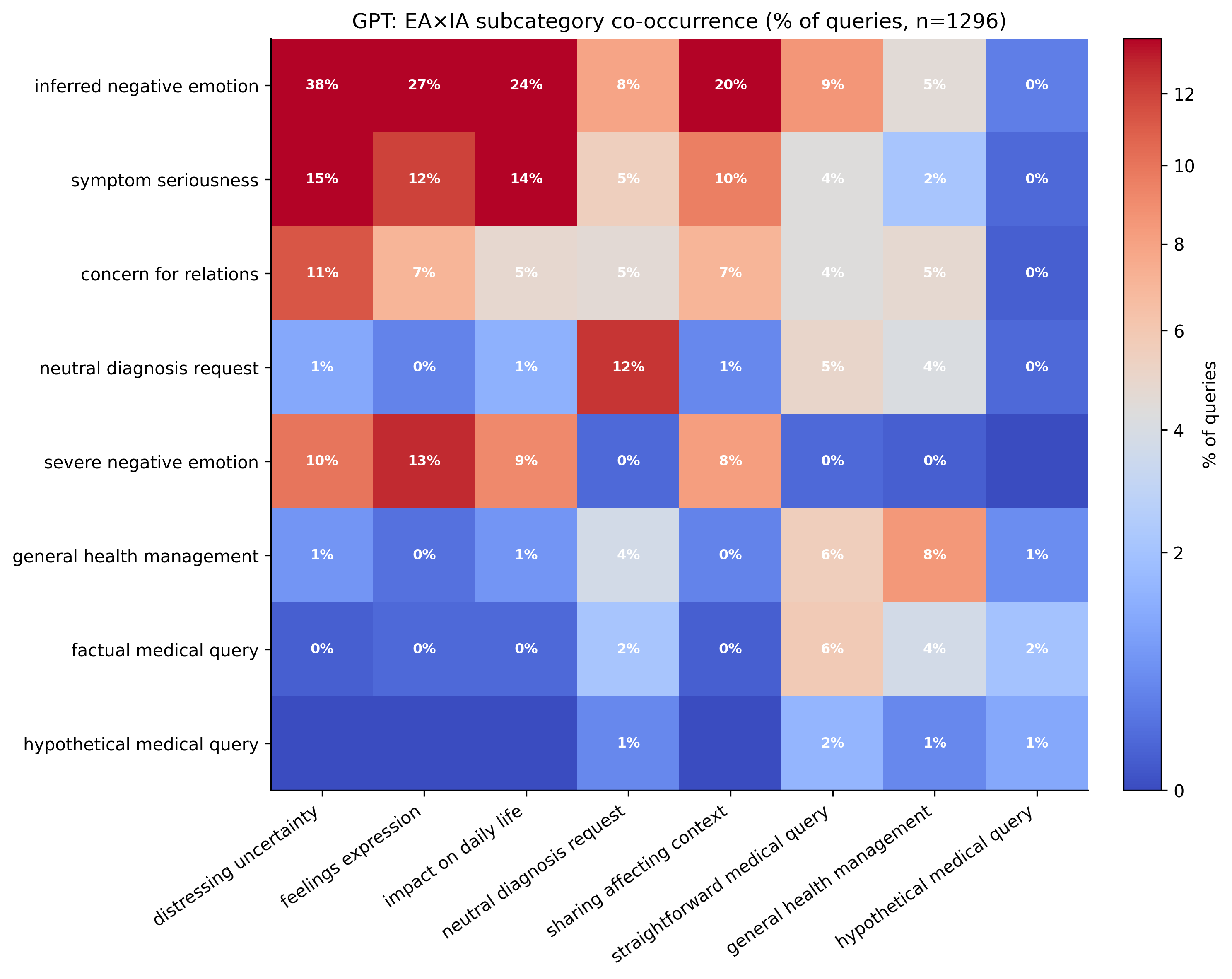}
  \caption{GPT: EA$\times$IA co-occurrence (multi-label IA/EA allowed).}
\end{subfigure}

\caption{EA$\times$IA subcategory co-occurrence (percent of queries, $N=1296$). Humans provide a single EA and IA subcategory per query, yielding sharper pairings; GPT may assign multiple subcategories per query, producing broader co-occurrence patterns.}
\label{fig:eaia-subcat-cooccur}
\end{figure*}

\subsection{Subcategory Prevalence and EA$\times$IA Co-occurrence (Humans vs.\ GPT)}
\label{app:subcat-prevalence-cooccur}

Table~\ref{tab:ea-prevalence-app} and Table~\ref{tab:ia-prevalence-app} report the prevalence of subcategories along with the \textbf{Applicable} or \textbf{Not Applicable} classification of each subcategory under the EAF. This makes it possible to assess the \emph{intuitive coherence} of the framework: subcategories that encode similar affective cues or uncertainty (Applicable) should align with each other in all dimensions, while informational or routine requests (Not Applicable) should be clustered separately.

Because each query receives exactly one subcategory from each human annotator, human counts sum to $N=1296$ per dimension. In contrast, GPT may assign multiple subcategories per query; thus GPT totals exceed $N$ (EA: 2100; IA: 2572), corresponding to an average of 1.62 EA subcategories and 1.98 IA subcategories per query.

The prevalence distributions show both stable structure and meaningful variability in how cues are operationalized. For EA, H1 frequently uses the Not Applicable subcategory \textit{Factual Medical Query} (31.8\%), while H2 rarely uses it (2.9\%) and instead assigns both Applicable and Not Applicable categories such as \textit{Concern for Relations} (20.6\%) and \textit{General Health Management} (19.4\%). A similar pattern appears in IA: H1 frequently assigns the Not Applicable subcategory \textit{Straightforward Medical Query} (33.3\%), whereas H2 more often assigns the Applicable subcategory \textit{Distressing Uncertainty} (40.8\%). GPT assigns Applicable EA categories at higher rates, especially \textit{Inferred Negative Emotion} (59.4\%) and \textit{Symptom Seriousness} (27.5\%), reflecting broader recall for empathic-need cues. Across dimensions, GPT again assigns multiple IA cues per query and frequently marks Applicable interpretive cues (\textit{Distressing Uncertainty}, \textit{Feelings Expression}, \textit{Impact on Daily Life}, \textit{Sharing Affecting Context}), consistent with its broader recall of applicability signals.

Figure~\ref{fig:eaia-subcat-cooccur} further supports the framework’s \emph{intuitive coherence} when interpreted through the Applicable vs.\ Not Applicable split of subcategories (Tables~\ref{tab:ea-prevalence-app}--\ref{tab:ia-prevalence-app}). Not Applicable request-types tend to pair with Not Applicable interpretations (Not$\times$Not), while affective/uncertainty cues (Applicable) more often co-occur with Applicable interpretive signals (App$\times$App). Concretely, in the human heatmaps, EA \textit{Factual Medical Query} aligns strongly with IA \textit{Straightforward Medical Query} (H1: 30\%), and \textit{Neutral Diagnosis Request} co-occurs with its IA counterpart (H1: 19\%; H2: 9\%), both canonical Not$\times$Not pairings. In contrast, Applicable EA cues align with Applicable IA cues: \textit{Severe Negative Emotion} co-occurs with \textit{Feelings Expression} (H1: 15\%), and \textit{Concern for Relations} frequently pairs with \textit{Distressing Uncertainty} (H2: 16\%), reflecting clinically intuitive links between expressed (or inferred) affect and corresponding interpretive needs. GPT exhibits broader App$\times$App cross-pairings, most prominently EA \textit{Inferred Negative Emotion} co-occurring with multiple Applicable IA cues including \textit{Distressing Uncertainty} (38\%), \textit{Feelings Expression} (27\%), \textit{Impact on Daily Life} (24\%), and \textit{Sharing Affecting Context} (20\%), which is expected given GPT’s multi-label rationale annotations. Overall, these structured pairings indicate that cross-dimensional co-occurrence is not arbitrary: it aligns with the EAF’s applicability semantics while also highlighting how multi-cue rationales (GPT) differ from single-label human assignments.

Additionally, these prevalence and co-occurrence results provide actionable signals for refining the EAF and improving annotation practice. First, the strong Not$\times$Not and App$\times$App structure suggests the framework’s applicability split is broadly coherent, but the sharp annotator skews in how “routine/informational” vs.\ “affective/uncertainty” cues are operationalized (e.g., heavier use of \textit{Factual Medical Query} and \textit{Straightforward Medical Query} versus greater use of \textit{Concern for Relations} and \textit{Distressing Uncertainty}) highlight subcategories that may be very broad or boundary-sensitive. These are prime candidates for guideline refinement (clearer decision rules, additional contrastive examples, or merging/splitting categories), while consistently rare categories can be reconsidered for consolidation if they contribute limited discriminative value. Second, the co-occurrence matrices can be leveraged to diagnose systematic annotation patterns and potential drift: stable, clinically intuitive pairings (e.g., \textit{Severe Negative Emotion} with \textit{Feelings Expression}) indicate consistent interpretation, whereas unexpected or diffuse pairings can reveal where annotators may be diverging due to inconsistent application rather than subjectivity, and targeted review could mitigate those differences. In this sense, co-occurrence structure is not only a validation of the EAF semantics, but also a practical tool for targeted adjudication, annotator training, and future modeling choices.

\begin{figure*}[t]
\centering

\begin{subfigure}[t]{0.95\textwidth}
  \centering
  \includegraphics[width=\linewidth]{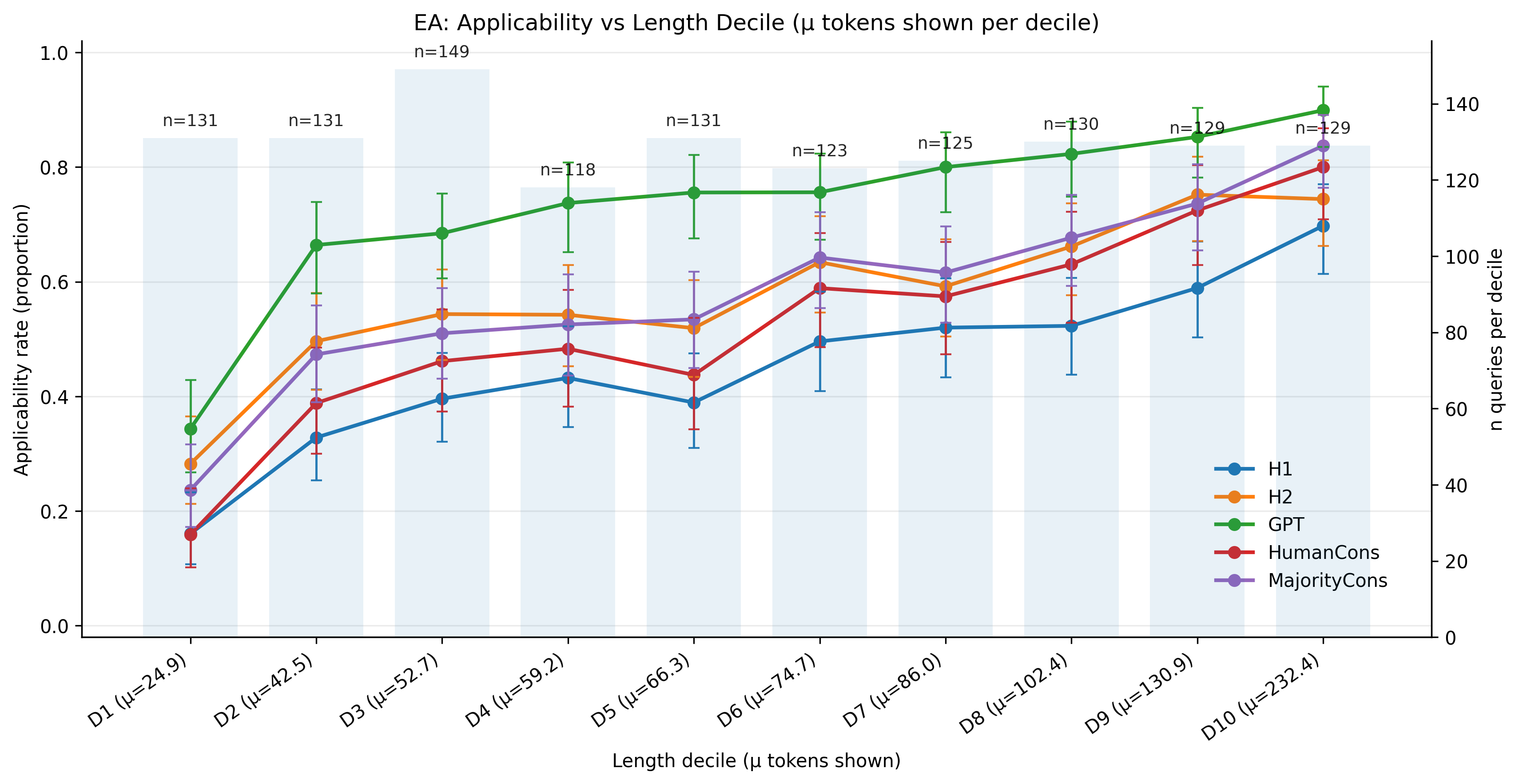}
  \caption{EA: Applicability vs.\ token-length decile ($\mu$ tokens shown per decile; error bars are confidence intervals).}
  \label{fig:ea-length-deciles}
\end{subfigure}

\vspace{0.8em}

\begin{subfigure}[t]{0.95\textwidth}
  \centering
  \includegraphics[width=\linewidth]{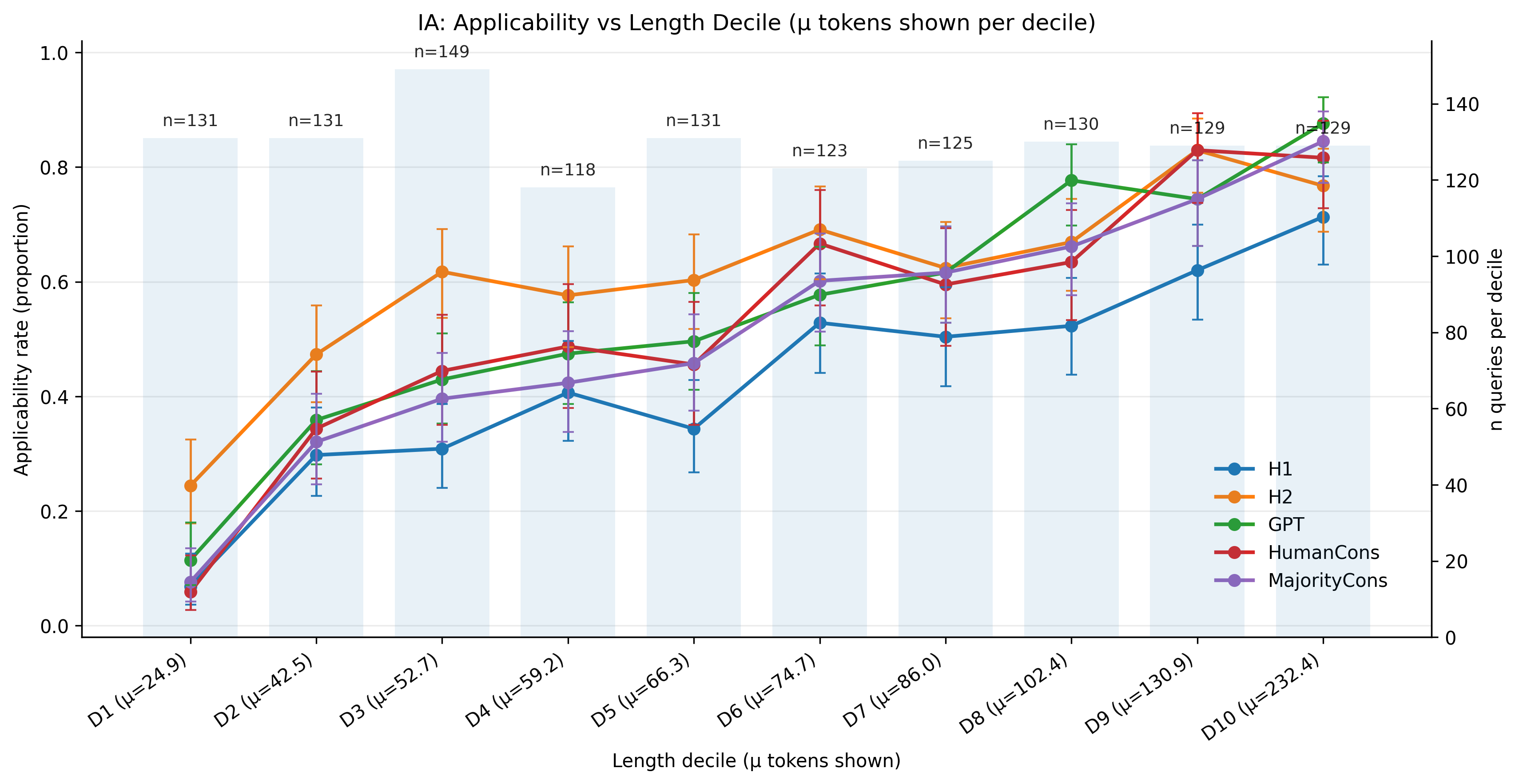}
  \caption{IA: Applicability vs.\ token-length decile ($\mu$ tokens shown per decile; error bars are confidence intervals).}
  \label{fig:ia-length-deciles}
\end{subfigure}

\caption{Applicability rates by query-length decile. Bars indicate the number of queries per decile ($n$). Error bars denote confidence intervals.}
\label{fig:length-effects}
\end{figure*}

\subsection{Length Effects on Applicability (with Confidence Intervals)}
\label{app:length-effects}
To assess whether query length is associated with applicability judgments, we stratify items into token-length deciles and plot EA/IA applicability rates for each label source (HA1, HA2, GPT, HumanCons, MajorityCons) (Figure \ref{fig:length-effects}). The x-axis reports the mean token length per decile ($\mu$), and we annotate the number of items per decile ($n$). Error bars denote confidence intervals for each applicability estimate.

Across both dimensions, applicability increases with length: shorter queries (D1--D2) receive substantially lower applicability rates, while longer queries (D9--D10) show consistently higher applicability across all sources. Although absolute rates differ by source (e.g., GPT tends to assign applicability more frequently than humans), the upward trend is shared, suggesting that length (and the additional context typically present in longer queries) is systematically associated with applicability rather than reflecting annotator idiosyncrasies alone.

\begin{figure*}[t]
\centering

\begin{subfigure}[t]{0.95\textwidth}
  \centering
  \includegraphics[width=\linewidth]{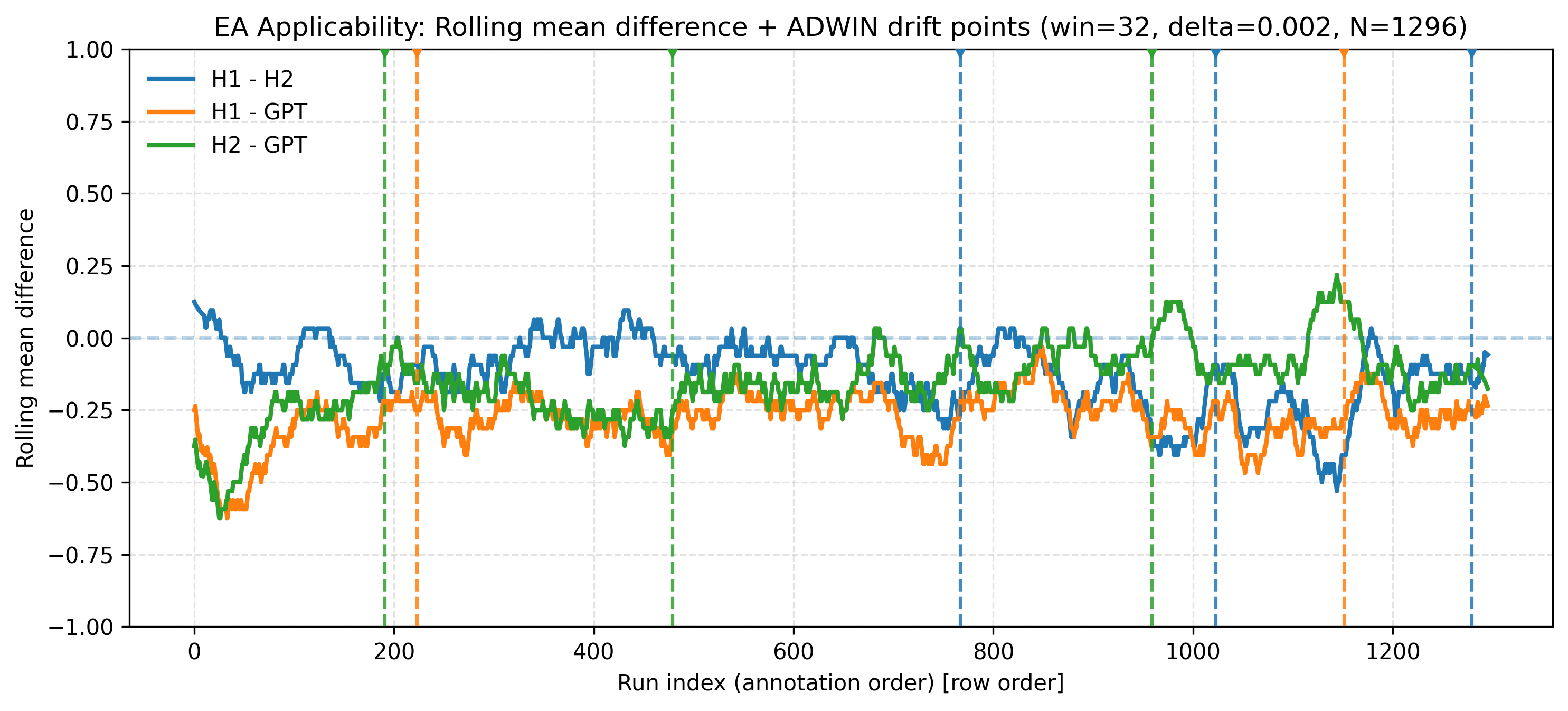}
  \caption{EA: rolling mean pairwise difference + ADWIN drift points (win=32, $\delta=0.002$).}
  \label{fig:drift-ea}
\end{subfigure}

\vspace{0.8em}

\begin{subfigure}[t]{0.95\textwidth}
  \centering
  \includegraphics[width=\linewidth]{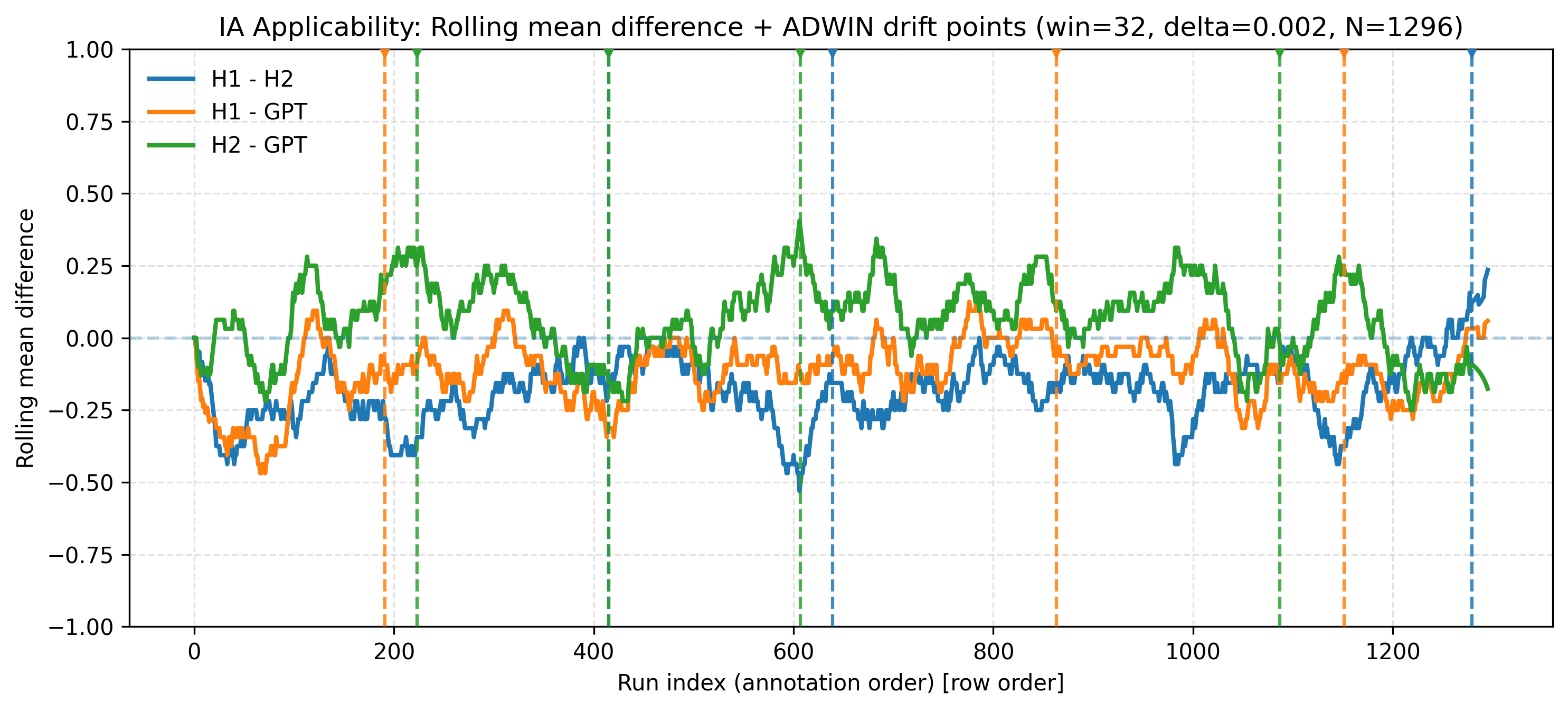}
  \caption{IA: rolling mean pairwise difference + ADWIN drift points (win=32, $\delta=0.002$).}
  \label{fig:drift-ia}
\end{subfigure}

\caption{Run-order drift diagnostics using rolling mean differences in applicability rates and ADWIN change detection. Vertical dashed lines indicate detected change points.}
\label{fig:drift-checks}
\end{figure*}

\subsection{Run-order Drift Checks}
\label{app:drift-checks}
Finally, we evaluate whether labeling behavior drifts over the annotation sequence (e.g., fatigue or order effects). We plot the rolling mean of pairwise differences in applicability rates (window size $=32$) across run index for each pair (H1--H2, H1--GPT, H2--GPT), and apply ADWIN (ADaptive WINdowing) change detection ($\delta=0.002$) to flag potential change points (Figure \ref{fig:drift-checks}).

Across $N=1296$ items, ADWIN identifies only a small number of localized change points in each rolling-difference stream (2--4 per pair for both EA and IA). Each detection reflects a local shift over the surrounding $\sim$32-item neighborhood rather than a single item. Overall, the sparse detections and the absence of sustained shifts in the rolling trajectories suggest broadly stable annotation behavior over the run, with only occasional local fluctuations in pairwise disagreement.

\subsection{Match vs.\ Miss by subcategory}
\label{app:match-miss}

\begin{figure*}[t]
    \centering
    \includegraphics[width=\textwidth]{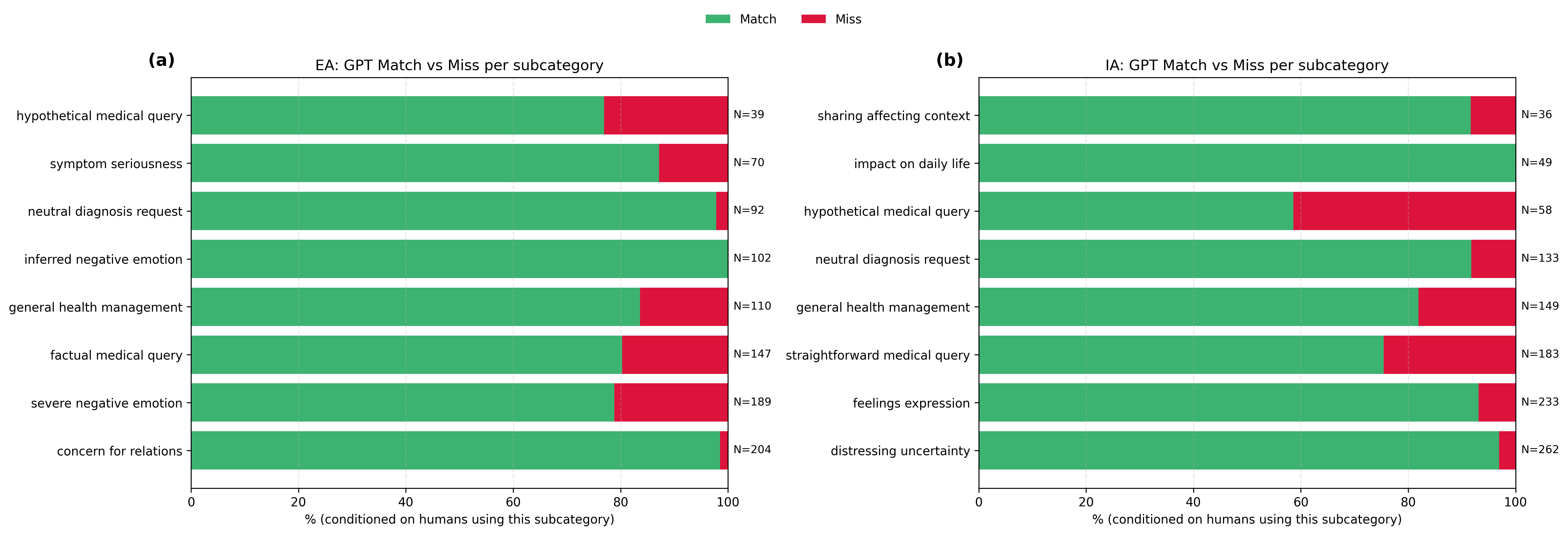}
    \caption{\textbf{Rationale Match vs.\ Miss by subcategory}.
    Stacked bars show, for each subcategory, the percentage of \textbf{Match} (green) vs.\ \textbf{Miss} (red) between GPT and the human-selected subcategory rationale, computed only on queries where GPT agrees with the human consensus applicability label. The $N$ label denotes the number of times the subcategory appears in the human rationale set within this agreement subset. Panels show \textbf{(a)} EA and \textbf{(b)} IA.}
    \label{fig:match_miss_by_subcat}
\end{figure*}

In this section, we quantify subcategory-level rationale overlap in cases where GPT agrees with the human consensus label (Applicable / Not Applicable) for the given dimension. For each query, each human provides one best-fit subcategory; the human rationale set is therefore either a singleton (both humans chose the same subcategory) or a size-two set (humans chose different subcategories). We then compare this human set to GPT's provided subcategories as rationales for labeling the query as Applicable or not. For each human-selected subcategory occurrence, we count a \textbf{Match} if GPT includes that same subcategory, and a \textbf{Miss} otherwise. Figure~\ref{fig:match_miss_by_subcat} reports, for each subcategory, the \% Match vs.\ \% Miss, along with the total number of such occurrences ($N$).

Across both EA and IA, match rates are generally high, indicating that when GPT agrees with humans on the binary applicability label, it often also identifies the same underlying rationale cues. However, several subcategories exhibit  higher miss rates, suggesting systematic differences in how GPT justifies an agreed-upon label.

For \textbf{EA} (Figure~\ref{fig:match_miss_by_subcat}a), GPT shows near-complete alignment for \emph{Concern for Relations} ($N=204$) and \emph{Inferred Negative Emotion} ($N=102$), and very strong alignment for \emph{Neutral Diagnosis Request} ($N=92$) and \emph{Symptom Seriousness} ($N=70$). In contrast, \emph{Hypothetical Medical Query} exhibits the largest miss portion ($N=39$), and both \emph{Severe Negative Emotion} ($N=189$) and \emph{Factual Medical Query} ($N=147$) show notable misses, indicating that GPT sometimes agrees on EA applicability while grounding its justification in different cues than the humans.

For \textbf{IA} (Figure~\ref{fig:match_miss_by_subcat}b), alignment is strongest for \emph{Distressing Uncertainty} ($N=262$) and \emph{Feelings Expression} ($N=233$), and is also high for \emph{Neutral Diagnosis Request} ($N=133$). \emph{Impact on Daily Life} shows a perfect match in this subset ($N=49$), though this category is comparatively smaller. The most challenging IA category is again \emph{Hypothetical Medical Query} ($N=58$), which has the largest miss fraction; \emph{Straightforward Medical Query} ($N=183$) also shows a higher miss rate than the more affective/uncertainty categories. 


Together, these patterns indicate substantial rationale overlap when GPT agrees with the human consensus applicability label, but alignment varies by subcategory. Misses occur not only for hypothetical and some informational categories, but also for certain affective categories (e.g., \emph{Severe Negative Emotion}). This suggests that mismatches reflect less a single cue-type split and more the prevalence of multi-cue queries, where multiple plausible rationales can support the same applicability judgment.

\end{document}